\documentclass[lettersize,journal]{IEEEtran}
\usepackage{amsmath,amsfonts}
\usepackage{algorithmic}
\usepackage{algorithm}
\usepackage{array}
\usepackage[caption=false,font=normalsize,labelfont=sf,textfont=sf]{subfig}
\usepackage{textcomp}
\usepackage{stfloats}
\usepackage{url}
\usepackage{verbatim}
\usepackage{graphicx}
\usepackage{cite}
\usepackage[table]{xcolor}
\usepackage{booktabs}
\usepackage{multirow}
\usepackage{microtype}
\usepackage{tabularx}
\usepackage{multirow}
\usepackage{siunitx}
\newcolumntype{C}{>{\centering\arraybackslash}X} 
\begin{document}

\title{Point-to-Mask: From Arbitrary Point Annotations to Mask-Level Infrared Small Target Detection}

\author{Weihua Gao, Wenlong Niu, Jie Tang, Man Yang, Jiafeng Zhang, Xiaodong Peng
\thanks{Manuscript received ; revised . (Corresponding author: Wenlong Niu.)}%
\thanks{This work was supported by the Civil Aerospace Pre-research Project under Grant D040103. All authors are with the Key Laboratory of Electronics and Information Technology for Space Systems, National Space Science Center, Chinese Academy of Sciences, Beijing 100190, China. Weihua Gao is also with the School of Computer Science and Technology, University of Chinese Academy of Sciences, Beijing 100049, China. (e-mail: gaoweihua22@mails.ucas.ac.cn; niuwenlong@nssc.ac.cn).}
}

\markboth{Journal of \LaTeX\ Class Files,~Vol.~14, No.~8, August~2021}%
{Shell \MakeLowercase{\textit{et al.}}: A Sample Article Using IEEEtran.cls for IEEE Journals}

\maketitle
\begin{abstract}
Infrared small target detection (IRSTD) methods predominantly formulate the task as pixel-level segmentation, which requires costly dense annotations and is not well suited to tiny targets with weak texture and ambiguous boundaries. To address this issue, we propose Point-to-Mask, a framework that bridges low-cost point supervision and mask-level detection through two components: a Physics-driven Adaptive Mask Generation (PAMG) module that converts point annotations into compact target masks and geometric cues, and a lightweight Radius-aware Point Regression Network (RPR-Net) that reformulates IRSTD as target center localization and effective radius regression using spatiotemporal motion cues. 
The two modules form a closed loop: PAMG generates pseudo masks and geometric supervision during training, while the geometric predictions of RPR-Net are fed back to PAMG for pixel-level mask recovery during inference. To facilitate systematic evaluation, we further construct SIRSTD-Pixel, a sequential dataset with refined pixel-level annotations. 
Experiments show that the proposed framework achieves strong pseudo-label quality, high detection accuracy, and efficient inference, approaching full-supervision performance under point-supervised settings with substantially lower annotation cost. Code and datasets will be available at: https://github.com/GaoScience/point-to-mask.
\end{abstract}
\begin{IEEEkeywords}
Infrared small target detection, point supervision, pseudo-mask generation, spatiotemporal modeling, weak supervision.
\end{IEEEkeywords}

\section{Introduction}

\IEEEPARstart{I}{nfrared} small target detection (IRSTD) plays an important role in applications such as long-distance early warning, low-altitude surveillance, and maritime search and rescue \cite{luo2025spatialtemporal, liu2024infrared,teutsch2010classification,rawat2020review,ding2025improving}. However, constrained by long-distance imaging conditions and infrared imaging mechanisms, targets typically occupy only a few pixels in the image, exhibiting physical characteristics such as weak textures, low signal-to-noise ratios (SNR), and severe boundary blurring. Complex background clutter, structured high-brightness interference, and intense imaging noise further increase the detection difficulty. In recent years, deep learning has improved the performance of IRSTD. Nevertheless, mainstream approaches predominantly model the task as a dense pixel-wise prediction or segmentation problem \cite{li2022dense, zhang2022isnet, sun2023receptivefield, yue2025sdsnet,guo2026multiscale}. 
Although this paradigm has improved detection accuracy, it still exhibits two limitations when dealing with extremely sparse foreground objects such as infrared small targets. First, the targets usually occupy only a tiny portion of the image, meaning that segmentation-based deep networks must devote substantial computation to suppressing background responses and often require long training iterations to converge, which is unfavorable for generalization. Second, these methods rely on expensive pixel-level annotations, which limits their scalability and practical deployment in large-scale real-world scenarios.

To reduce modeling and annotation costs, some studies have borrowed insights from generic object detection by introducing bounding box-based supervision \cite{yang2024eflnet, tong2024sttrans, chen2024sstnet, zhu2024tmp, duan2024tripledomain}. Other efforts have further adopted single point annotations as a more lightweight supervision signal \cite{li2023monte, kou2024mcgc}. However, both paradigms exhibit limitations in IRSTD. 
Due to the optical point spread function (PSF), infrared small targets usually appear as compact yet blurred thermal responses rather than objects with clear geometric boundaries \cite{zhang2024irsam}. Under this imaging characteristic, rigid rectangular boxes are often unable to tightly describe the target support and may introduce considerable background redundancy, which is especially problematic in low SNR scenarios. 
Point supervision provides a more economical form of annotation and is therefore increasingly explored in IRSTD. However, many existing approaches still rely on additional assumptions, such as predefined scale priors, highly accurate center clicks, or intermediate feature representations whose stability varies across scenes and training stages \cite{he2025hybrid, ying2023mapping, yu2025from}. When the annotated point deviates from the true target center, the recovered pseudo masks may become unstable, which in turn affects the reliability of subsequent detection models. 
Consequently, the key difficulty is not merely adopting weaker supervision, but how to convert sparse point annotations into stable and physically meaningful target support that can serve as effective supervisory signals for downstream learning \cite{ni2025pointtopoint}.

Rethinking the imaging characteristics of infrared small targets suggests that the most informative cues often lie not in complex contours, but in the target center and its effective spatial support. Compared with coarse bounding-box representations or computationally expensive dense segmentation, a more suitable modeling choice may be the compact geometric support of the target. 
This is because infrared small targets typically appear as compact thermal spots influenced by the point spread effect, whose responses are close to isotropic or compact diffusion patterns. Under such conditions, regressing width and height with bounding boxes often fails to accurately reflect the underlying physical shape, and these scale parameters are not always necessary, potentially introducing additional prediction uncertainty. In contrast, representing the target using its center location and effective radius better aligns with the imaging characteristics of infrared small targets, while preserving target energy and reducing background redundancy.

To instantiate this concept, we construct a detection framework composed of two complementary components. The first component is the Physics-driven Adaptive Mask Generation (PAMG) algorithm, which generates reliable pseudo masks from sparse point supervision. By exploiting the physical characteristics of infrared small targets, PAMG can adaptively grow a compact target mask starting from an arbitrary point annotation within the target region, and further extract geometric supervision signals such as the centroid and the equivalent radius.
Building upon this, we design the lightweight Radius-aware Point Regression Network (RPR-Net), reformulating the detection task as center localization and effective radius regression. Rather than relying on full-image segmentation, RPR-Net directly regresses the target center and effective radius, and incorporates spatiotemporal kinematic cues to compensate for the scarcity of single-frame appearance information.

This design also forms a consistent training--inference pipeline. During the training phase, single point annotations are used to generate pseudo-masks via PAMG, from which compact geometric supervision signals are extracted to train the network. During the inference phase, the position and radius predicted by RPR-Net can be fed back into PAMG to recover pixel-level masks when finer target support is needed. This mechanism provides a flexible configuration for IRSTD: the network outputs can serve as detection results when only target location and scale are required, while masks can be further recovered for tasks that require target shapes.

Finally, given that existing datasets providing precise pixel-level annotations are predominantly single-frame \cite{jiang2023antiuav, li2022dense, zhang2022isnet} and often suffer from subjective annotation bias, limited scale, and restricted scene coverage, we develop a specialized annotation tool, Label-IRST, based on PAMG, and construct SIRSTD-Pixel, a spatiotemporal benchmark under real-world scenarios. Using this dataset, we evaluate the proposed framework in terms of pseudo-label generation quality, full/point-supervised detection performance, component ablation, and efficiency analysis. Experimental results suggest that the framework reduces the dependence of point supervision on precise priors while maintaining competitive detection performance, offering a reasonable balance between inference efficiency and shape awareness.

The main contributions of this work are summarized as follows:
\begin{enumerate}
    \item We propose a Point-to-Mask detection paradigm for IRSTD. Guided by the physical imaging characteristics of infrared small targets, we represent the detection object with the target center and compact spatial support, linking low-cost supervision with mask-level detection.
    \item We design a closed-loop framework consisting of PAMG and RPR-Net. PAMG converts arbitrary point annotations into compact target masks and geometric supervision, while RPR-Net incorporates spatiotemporal cues to localize target centers and regress effective radii; during inference, its geometric predictions can be fed back to PAMG for pixel-level mask recovery.
    \item We develop the annotation tool Label-IRST and construct the spatiotemporal benchmark SIRSTD-Pixel. Experiments show that the proposed framework achieves strong pseudo-label quality, detection performance, and inference efficiency, while remaining close to full supervision under point-supervised settings.
\end{enumerate}

\section{Related Work}

\subsection{Fully-Supervised Infrared Small Target Detection}

Early infrared small target detection methods mainly relied on traditional techniques. These approaches typically exploit the local saliency of targets by enhancing target responses while suppressing background interference, such as the classical IPI and LCM algorithms \cite{gao2013infrared,liu2023singleframe,chen2013a,li2023sparse}. With the development of deep learning, significant progress has been achieved in IRSTD, and the dominant research paradigm has gradually shifted toward treating the task as a pixel-level supervised image segmentation problem.

Within this paradigm, early convolutional neural network (CNN)-based approaches mainly improved the separability between targets and background through careful feature design. Typical strategies include cross-layer feature fusion, attention mechanisms, and local contrast modeling to enhance target responses. Representative methods include ACM \cite{dai2021asymmetric}, ALCNet \cite{dai2021attentional}, and DNANet \cite{li2022dense}. 
Subsequently, more specialized architectures were proposed to address the morphological characteristics of infrared small targets and the challenges posed by complex backgrounds. For example, ISNet \cite{zhang2022isnet} enhances detection by introducing a target shape reconstruction mechanism, while shape-biased representation learning \cite{lin2024learning} further encourages networks to capture structural cues of small targets. MSHNet \cite{liu2024infrared} improves sensitivity to target scale and location through a tailored loss design. 
In addition, Transformer-based context-aware models have been introduced into IRSTD to capture longer-range dependencies and further improve detection performance in complex backgrounds \cite{liu2023infrared,yuan2024sctransnet}.

Beyond appearance modeling in single frames, temporal modeling has recently become an important direction for sequence-based IRSTD. These methods exploit motion consistency across consecutive frames to construct more robust spatiotemporal representations. For instance, DTUM \cite{li2025directioncoded} introduces direction-encoded temporal difference information into a U-shaped architecture, while LMAFormer integrates motion-aware spatiotemporal attention with a multi-scale Transformer encoder to jointly model target motion and background dynamics \cite{huang2024lmaformer}.

Taken together, whether in single-frame or sequential detection scenarios, most existing approaches still rely on pixel-wise masks and follow a dense segmentation paradigm. For extremely small, sparse, and blurry infrared targets, this formulation is effective but may not always be the most appropriate choice in terms of annotation cost and representation compactness, leaving room for more geometry-oriented target representations.

\subsection{Weakly-Supervised Detection: From Box to Point}

Compared with pixel-level segmentation supervision, bounding-box-based detection methods generally require lower annotation cost. This line of work mainly follows the modeling paradigm of generic object detection by regressing target location and scale to perform object-level detection. In IRSTD, although single-frame box-based detectors such as EFLNet have been explored \cite{yang2024eflnet}, many existing methods adopt multi-frame inputs because box regression can be naturally combined with spatiotemporal feature extraction modules. For example, TMP, DBMSTN, and STME employ dual-branch architectures to separately extract appearance and motion features \cite{zhu2024tmp,li2025dbmstn,peng2025moving}; SSTNet captures implicit spatiotemporal information through cross-slice ConvLSTM modules \cite{chen2024sstnet}; and ST-Trans directly processes consecutive frames with spatiotemporal transformation modules to output target bounding boxes \cite{tong2024sttrans}. In this sense, box supervision offers a practical compromise between annotation cost and object-level detection requirements.

However, bounding-box representations are not perfectly suited for infrared small targets. Unlike natural objects that usually occupy large spatial regions, infrared small targets often appear as compact thermal spots consisting of only a few pixels, with inherently ambiguous boundaries. Loose bounding boxes tend to introduce substantial background interference, while overly tight boxes may fail to cover the effective spatial support of the target. Therefore, although box supervision is more economical than pixel-level supervision, it still has clear limitations in accurately representing the true support region of infrared small targets.

In contrast, modeling infrared small targets as points better reflects their compact characteristics. Direct point regression methods further simplify the representation. For example, P2P directly models targets as points and performs detection through point regression networks \cite{ni2025pointtopoint}. However, representing targets solely as points inevitably loses important information such as target scale and spatial support. Consequently, recent studies have increasingly focused on point-supervised methods that recover target masks from point prompts. LESPS observes that networks often respond to target shapes before regressing the target point and therefore attempts to reconstruct masks from intermediate responses \cite{ying2023mapping}; PAL further stabilizes label evolution through progressive regularization \cite{yu2025from}. Another group of methods adopts more explicit pseudo-mask generation strategies. For example, MCLC, COM, and MCGC progressively recover pseudo masks from point prompts through clustering, contour evolution, or region-growing strategies \cite{li2023monte,li2024a,kou2024mcgc}, while HMG combines handcrafted and learning-based mask generation mechanisms \cite{he2025hybrid}. In addition, the Segment Anything Model (SAM) also demonstrates the flexibility of point prompts in segmentation tasks, although its general visual priors are not specifically designed for recovering the compact support of infrared small targets \cite{kirillov2023segment}.

Despite these encouraging results, notable limitations remain in point-supervised IRSTD. On the one hand, some approaches rely on intermediate network responses, which are often unstable or prone to excessive expansion \cite{ying2023mapping,yu2025from}. On the other hand, many methods still depend on additional prior assumptions, such as requiring the point prompt to be near the target center, using predefined cropping windows, or assuming specific scale ranges \cite{li2023monte,kou2024mcgc,he2025hybrid}. As a result, reliably recovering a compact target support region that is consistent with the underlying physical response from an arbitrary point within the target remains an open problem.

\subsection{Datasets and Benchmarks for IRSTD}

The development of datasets has played a crucial role in advancing infrared small target detection from traditional approaches to deep learning methods. In recent years, datasets such as SIRST, NUDT-SIRST, and IRSTD-1K have been proposed, providing relatively standardized benchmarks for pixel-level detection in real infrared imagery and promoting the development of single-frame IRSTD methods \cite{dai2021asymmetric,li2022dense,zhang2022isnet}. However, these datasets are typically limited in scale and mainly support single-frame image analysis, making them less suitable for evaluating temporal consistency and detection behavior in continuous scenarios.

To better reflect real-world search, tracking, and continuous observation scenarios, several sequence datasets have subsequently been introduced. For example, the DAUB dataset provides annotated infrared image sequences for sequence-based research, although its scale and diversity remain limited \cite{hui2019a}. Larger video resources such as SIRSTD and Anti-UAV contain richer motion patterns and more realistic application scenarios \cite{tong2024sttrans,jiang2023antiuav}. In addition, IRDST is also a widely used dataset, but its construction emphasizes scene diversity and sequence organization rather than strictly consistent pixel-level annotations across frames \cite{sun2023receptivefield}.

From a benchmarking perspective, existing sequence datasets are often designed primarily for detection or tracking tasks and generally lack high-precision, cross-frame-consistent pixel-level annotations. Consequently, despite the growing interest in point supervision and sequence modeling, there is still a lack of a unified benchmark that simultaneously supports point-supervised pseudo-mask evaluation, sequence detection, and pixel-level recovery analysis.

\section{Methodology}
\begin{figure*}[t]
    \centering
    \includegraphics[width=1\linewidth]{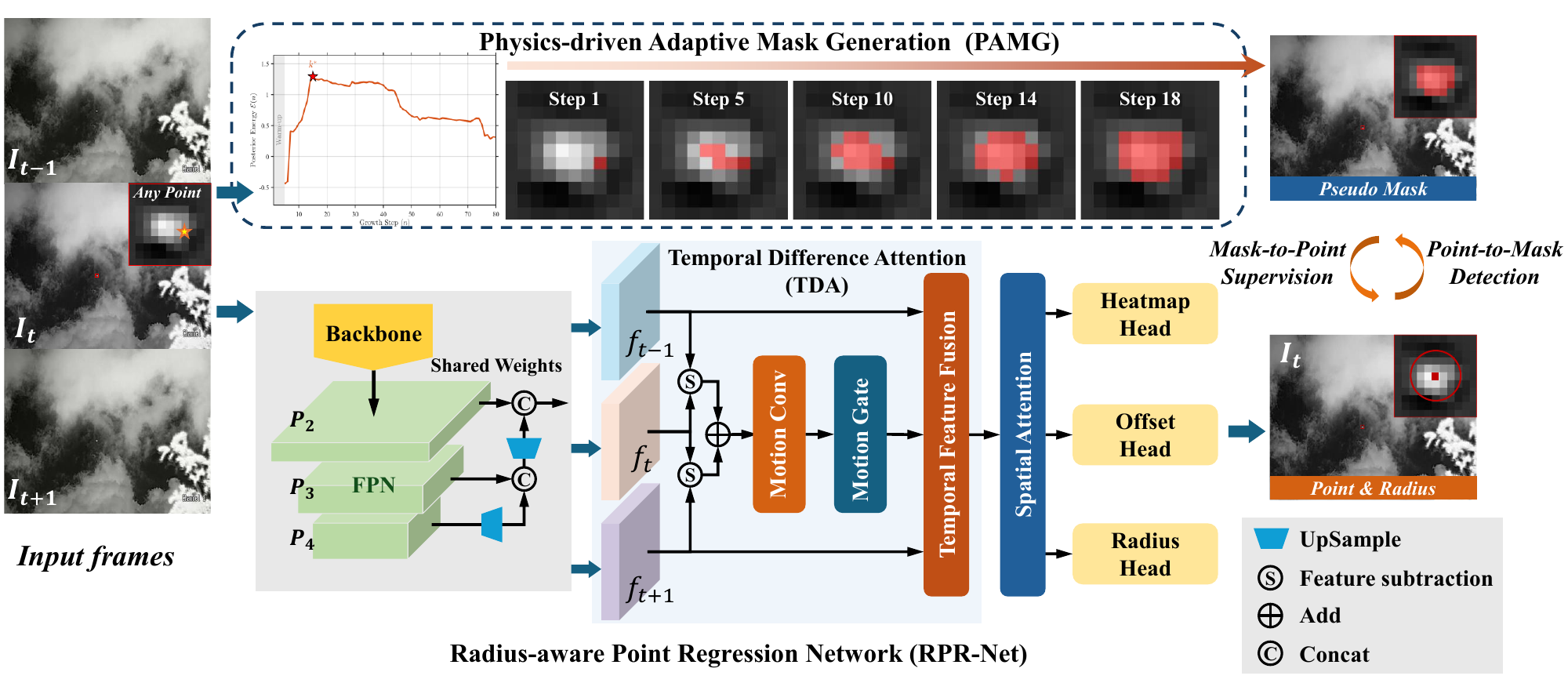} 
    \caption{Schematic illustration of the proposed framework. It integrates a Data Stream (top) and a Network Stream (bottom). The PAMG algorithm leverages physics priors to evolve random points into pixel-level masks, generating ground truth for the RPR-Net. The RPR-Net then predicts the target geometry parameters from input sequences using a lightweight point regression architecture. This establishes a robust "Mask-to-Point Supervision" and "Point-to-Mask Detection" cycle.}
    \label{fig:overview}
\end{figure*}

This section presents the proposed Point-to-Mask framework for infrared small target detection. As shown in Fig.~\ref{fig:overview}, the framework consists of two components: the PAMG algorithm and the RPR-Net. PAMG converts a point annotation into a compact target mask and provides geometric supervision in the form of target centroids and effective radii. Based on these supervisory signals, RPR-Net predicts the target geometry parameters directly from image sequences using a lightweight point-regression architecture.

The two components are used differently during training and inference. During training, the centroids and radii extracted from PAMG-generated masks are used to supervise RPR-Net, reducing the ambiguity introduced by random point annotations. During inference, RPR-Net produces efficient geometric predictions, and the predicted radius can be further used as a scale cue for PAMG when finer target support recovery is needed.

The following subsections describe PAMG in Section~\ref{subsec:PAMG} and RPR-Net in Section~\ref{subsec:network}.

\subsection{Physics-driven Adaptive Mask Generation}
\label{subsec:PAMG}

To address the issues where existing point-supervised methods are sensitive to initial point locations and require size priors or rely on unstable intermediate network features, we model the mask generation process as a Bayesian Maximum A Posteriori (MAP) estimation problem. Leveraging the physical characteristics of infrared small targets, our proposed Physics-driven Adaptive Mask Generation (PAMG) algorithm adaptively evolves from an arbitrary seed point to the optimal target region by optimizing a physics-informed energy function. The specific formulation and optimization of PAMG are detailed below.

\subsubsection{Problem Formulation via MAP Estimation}
Let $D$ denote the observed infrared image data, and $S_n$ represent a candidate target region consisting of $n$ pixels, which is iteratively grown from an initial seed point. Our objective is to identify the optimal region $S^*$ that maximizes the posterior probability $P(S_n | D)$ given the observation $D$. According to Bayes' theorem, this can be formulated as:
\begin{equation}
S^* = \underset{S_n}{\arg\max} \ P(S_n | D) = \underset{S_n}{\arg\max} \ \frac{P(D | S_n) \cdot P(S_n)}{P(D)}
\end{equation}

Since the evidence factor $P(D)$ is constant for a given image, maximizing the posterior is equivalent to maximizing its logarithm. This transformation converts multiplicative probability terms into an additive energy formulation, which facilitates joint modeling and improves numerical stability. The optimization objective is therefore defined as maximizing the following energy function $\mathcal{E}(S_n)$:

\begin{equation}
\mathcal{E}(S_n) \propto \ln P(D | S_n) + \ln P(S_n) = \mathcal{L}_{data}(S_n) + \mathcal{L}_{prior}(S_n)
\end{equation}
where $\mathcal{L}_{data}$ serves as the data likelihood term quantifying the target's thermal response, and $\mathcal{L}_{prior}$ acts as the prior term imposing physical constraints on the geometric shape and size.

\subsubsection{Construction of Physics-Informed Energy Function}
To solve the MAP problem without introducing heuristic or artificial constraints, we explicitly construct the energy components strictly based on three physical phenomena of infrared small targets: \textit{Local Saliency}, \textit{Internal Homogeneity}, and the \textit{Point Spread Effect}.

\begin{itemize}
    \item Likelihood Term: Homogeneity-Weighted SNR \\
    Infrared small targets typically exhibit prominent thermal signatures (\textit{Local Saliency}) but lack rich texture, manifesting as smooth regions (\textit{Internal Homogeneity}). To capture this "bright and uniform" physical nature, we model the data likelihood $P(D|S_n)$ as a modified t-statistic, termed Homogeneity-Weighted SNR (Hw-SNR):
    \begin{equation}
    \mathcal{L}_{data}(S_n) = \ln \left( \frac{\mu_{in} - \mu_{out}}{\sigma_{in} + \epsilon} \right)
    \end{equation}
    where $\mu_{in}$ and $\mu_{out}$ represent the mean intensity of the candidate region $S_n$ and its immediate background, respectively. The numerator $(\mu_{in} - \mu_{out})$ measures the effective signal contrast. The denominator $\sigma_{in}$ represents the internal standard deviation of the region, and $\epsilon$ is a small constant for numerical stability. 
    
    The introduction of $\sigma_{in}$ fundamentally distinguishes Hw-SNR from the traditional Signal-to-Clutter Ratio (SCR), which typically utilizes background deviation ($\sigma_{out}$). Here, we redefine "noise" as the structural uncertainty within the candidate region. By explicitly penalizing regions with drastic internal texture variations, Hw-SNR ensures that the algorithm prioritizes convergence on authentic, homogeneous thermal targets rather than highly-responsive but chaotic clutter.

    \item Dynamic Gain Term: Entropy Size Prior \\
    A realistic target is composed of a cluster of pixels rather than a single isolated point. During the nascent growth phase (when $n$ is very small), relying solely on the data likelihood may trap the algorithm at local noise peaks, leading to trivial solutions or growth stagnation. To provide early growth momentum, we introduce an entropy-driven size prior:
    \begin{equation}
    \mathcal{L}_{size}(S_n) = \ln(\ln n)
    \end{equation}
    The double-logarithmic form guarantees scale alignment with other energy terms. In the initial stage, this term provides a large gradient to push the region out of noise-induced local optima. As $n$ increases, its growth rapidly saturates, gracefully handing over the optimization dominance back to the data and geometric terms without forcing over-expansion.

    \item Penalty Term: Spatial Geometric Prior \\
    Relying exclusively on the likelihood term would cause the region to greedily swallow surrounding high-response pixels, resulting in severe background leakage. Due to the diffraction effects of the optical system (\textit{Point Spread Effect}), distant infrared targets manifest as blurred, isotropic Gaussian spots rather than elongated strips or irregular shapes. Correspondingly, we assume the spatial distribution of target pixels follows a 2D Gaussian penalty:
    \begin{equation}
    \mathcal{L}_{geo}(S_n) \propto - \frac{d_{max}^2}{2R_s^2}
    \end{equation}
    where $d_{max}$ denotes the maximum Euclidean distance from the current boundary pixels to the seed point, and $R_s$ is a spatial support parameter governing the tolerance of the geometric constraint. 
\end{itemize}

Integrating the likelihood, entropy size prior, and spatial geometric prior, the total physics-informed energy function is formulated as:
\begin{equation}
\mathcal{E}(S_n) = \ln(\ln n) + \ln \left( \frac{\mu_{in} - \mu_{out}}{\sigma_{in} + \epsilon} \right) - \frac{d_{max}^2}{2R_s^2} 
\label{equ:energy}
\end{equation}

\subsubsection{Optimization via Greedy Search}

As derived in the previous section, the objective of target segmentation is to find the optimal pixel set $S^*$ that maximizes the posterior energy function $\mathcal{E}(S_n)$. Since the search space of all possible pixel combinations is exponential, finding the global optimum is computationally intractable. However, under the assumptions of \textit{local saliency} and the \textit{point spread effect}, the target intensity typically exhibits a centralized distribution that decays toward the periphery. This physical property implies that selecting the neighbor with the highest intensity for each expansion step is the most effective path toward achieving maximum energy gain.

Therefore, we implement a deterministic optimization strategy leveraging a priority queue and greedy search to ensure that the candidate region is expanded consistently to maximize the energy function $\mathcal{E}(S_n)$. The complete optimization procedure is summarized in Algorithm \ref{alg:PAMG}.

\begin{algorithm}[t]
    \caption{Physics-driven Adaptive Mask Generation}
    \label{alg:PAMG}
    \begin{algorithmic}[1]
        \REQUIRE Image $I$, Seed point $p_{seed}$, Spatial support parameter $R_s$.
        \ENSURE Binary Mask $M$.
        
        \STATE \textbf{Step 1: Polarity Unification}
        \STATE Estimate local background median $b_{loc}$ around $p_{seed}$.
        \IF{$I(p_{seed}) < b_{loc}$}
        \STATE $I \leftarrow 1.0 - I$ \quad \COMMENT{Invert image for dark targets}
        \ENDIF
        
        \STATE \textbf{Step 2: Initialization with Seed Anchoring}
        \STATE Initialize Priority Queue $Q$.
        \STATE Push $p_{seed}$ into $Q$ with priority $-I(p_{seed})$.
        \STATE \textit{// Initialize running statistics for the candidate region:}
        \STATE $\mu_{in} \leftarrow I(p_{seed}), \sigma_{in} \leftarrow 0, n \leftarrow 1$. 
        \STATE $Path \leftarrow \emptyset$, $EnergyList \leftarrow \emptyset$.
        
        \STATE \textbf{Step 3: Greedy Growth with Warm-up}
        \WHILE{$Q \neq \emptyset$ \textbf{and} $n < \pi R_s^2$}
            \STATE Pop $p_{curr}$ with max intensity from $Q$.
            \STATE Append $p_{curr}$ to $Path$.
            
            \STATE \textit{// Update statistics:}
            \STATE $n \leftarrow n + 1$
            \STATE Update $\mu_{in}$ and $\sigma_{in}$ incrementally using $I(p_{curr})$.
            
            \STATE \textit{// Energy calculation with Warm-up constraint:}
            \IF{$n > 5$} 
                \STATE Calculate $\mathcal{E}(S_n)$ using Eq.~\ref{equ:energy}.
                \STATE Append $\mathcal{E}(S_n)$ to $EnergyList$.
            \ELSE
                \STATE Append $-\infty$ to $EnergyList$. \quad \COMMENT{Warm-up phase}
            \ENDIF
            
            \FOR{each neighbor $p_{next}$ of $p_{curr}$}
                \IF{$p_{next}$ not visited}
                    \STATE Push $p_{next}$ to $Q$, mark visited.
                \ENDIF
            \ENDFOR
        \ENDWHILE
        
        \STATE \textbf{Step 4: Mask Generation}
        \STATE Identify the optimal growth step: $k^* = \arg\max_{n} \{ \mathcal{E}(S_n) \}$.
        \STATE Construct the final binary mask $M$ using the first $k^*$ pixels in $Path$.
        \RETURN $M$

\end{algorithmic}

\end{algorithm}

To facilitate a comprehensive understanding of the operational mechanism of PAMG, we delineate four pivotal components that ensure its robustness and accuracy:

\begin{itemize}
    \item Unified Polarity Strategy: 
    While the energy function is derived assuming bright targets, dark targets also exhibit significant local contrast relative to the background, adhering to the \textit{local saliency} assumption. By comparing the seed intensity with the local background median $b_{loc}$, we invert the image intensity when necessary ($I' = 1 - I$). This preprocessing unifies the task as a peak-intensity search, significantly simplifying the generative logic for diverse target types.
    \item Robust Initialization and Statistical Warm-up: 
    The reliability of the Hw-SNR criterion heavily depends on the statistical significance of the internal standard deviation $\sigma_{in}$. However, during the nascent growth phase (e.g., $n < 5$), the sample size is insufficient to yield a stable estimate, often leading to numerical singularity or pseudo-energy peaks on single-pixel noise. To mitigate this "cold start" instability, we implement a dual-mechanism strategy:
    (1) Seed Anchoring: We pre-load the statistical accumulators with the seed point's properties, acting as a strong prior to anchor the initial growth.
    (2) Warm-up Constraint: We enforce a protective phase where energy evaluation is suspended until the region expands to a minimum cluster size.
    This combination ensures numerical stability and aligns with the physical prior that infrared targets, governed by the PSF, inherently occupy a non-negligible spatial support.
    
    \item Backtracking-based Mask Generation: 
    Unlike conventional methods that rely on rigid stopping thresholds, PAMG records the comprehensive growth sequence $\mathcal{S} = \{S_1, S_2, \dots, S_n\}$ alongside the corresponding posterior energy $\mathcal{E}(S_n)$, determining the final mask via a \textit{retrospective backtracking} mechanism. Due to the \textit{point spread effect}, the energy profile of infrared small targets often exhibits fluctuations near the boundary. Consequently, relying solely on early-stopping mechanisms may lead to entrapment in local optima. By analyzing the complete energy trajectory, we define the optimal growth step $k^*$ as the state achieving the global maximum of the posterior energy:
    \begin{equation}
    k^* = \arg\max_{n} { \mathcal{E}(S_n) }.
    \end{equation}
    This mechanism guarantees that the reconstructed mask $M$ represents the target region with the highest statistical significance. By isolating the global energy peak, the algorithm effectively suppresses false alarms induced by background leakage or edge diffusion.
    
\end{itemize}

\subsubsection{Theoretical Analysis on Mask Generation}
\label{sec:pamg_method}
This subsection analyzes the behavior of PAMG from three aspects: the growth dynamics under different seed locations, the role of the spatial support parameter $R_s$, and the existence of a finite optimum of the discrete energy sequence. These analyses explain why PAMG can generate stable supervisory masks under imperfect point annotations and ambiguous target boundaries.

\paragraph{Analysis of Growth Dynamics under Varying Initialization}
\begin{figure}[t]
    \centering
    \includegraphics[width=1\linewidth]{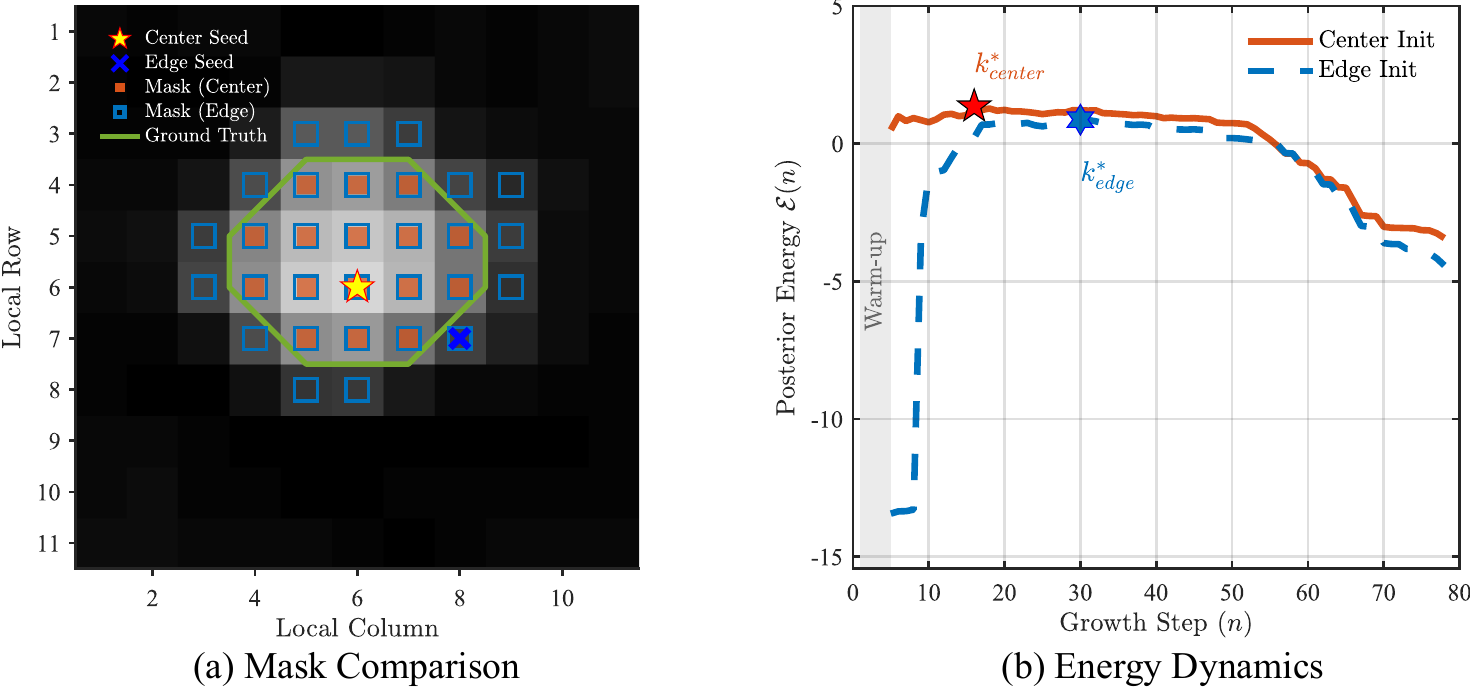}
    \caption{Visual analysis of PAMG robustness. (a) Comparison of masks generated from center ($\star$) and edge ($\times$) seeds, demonstrating that both cover the effective target core. (b) Posterior energy curves showing that despite an initial "climbing" lag for the edge seed, both trajectories converge to distinct but valid energy peaks after the warm-up phase.}
    \label{fig:initseed_change_fig}
\end{figure}

Ideally, the generated mask should be strictly invariant to the seed point location $p_{seed}$. However, due to the \textit{point spread effect}, infrared small targets exhibit fuzzy transition zones rather than sharp physical boundaries, making the attribution of boundary pixels inherently ambiguous. As a result, different initialization locations may induce minor variations in the final mask. This phenomenon, visualized in Fig. \ref{fig:initseed_change_fig}, is analyzed below through the lens of \textit{core consistency}. 
\begin{itemize}
    \item Center Initialization:
We first consider the case where $p_{seed}$ is the target center, aligned with the thermal centroid. Under this condition, the algorithm achieves the maximum Hw-SNR in the initial stage. Consequently, the growth process is relatively conservative, prioritizing the expansion of the high-intensity core region. As the region gradually extends into the fuzzy penumbra (i.e., the transition zone), the internal mean intensity $\mu_{in}$ decreases slightly. This causes the energy peak $k^* = \arg\max_{n} { \mathcal{E}(S_n) }$ to occur earlier, resulting in a compact mask $\mathcal{M}_{center}$ that tightly encloses the region of highest energy. 
    \item Edge Initialization:
Next, we consider the case where $p_{seed}$ is located at the target boundary. In this scenario, the algorithm undergoes an "anisotropic climbing" phase to progressively reach the target centroid. By the time the target is fully covered, the accumulated statistics differ slightly from the center-initialized case. The momentum gained during the climbing phase typically encourages the growth process to extend further into the transition region between the target and the background, thereby generating a slightly larger mask $\mathcal{M}_{edge}$.
\end{itemize}

Although $\mathcal{M}_{center} \neq \mathcal{M}_{edge}$ in terms of absolute pixel count, they satisfy the core containment property:
\begin{equation}
\mathcal{S}_{core} \subseteq \mathcal{M}_{center} \subseteq \mathcal{M}_{edge} \subseteq \mathcal{S}_{phy}
\end{equation}
where $\mathcal{S}_{core}$ represents the discriminative high-energy core, and $\mathcal{S}_{phy}$ represents the physical target extent considering all fuzzy boundary pixels.

Given that the primary objective of pseudo-label generation is to provide supervision on target presence and location, slight discrepancies in boundary definition are acceptable. Crucially, both masks cover the effective energy core, ensuring that different seed locations still provide valid supervisory signals. This also indicates that the final mask is not determined by initialization alone, but jointly by the growth path and the regularization imposed by the geometric prior. We therefore further examine the role of the spatial support parameter $R_s$.

\paragraph{Analysis of the Spatial Support Parameter $R_s$}
In our formulation, once the initial seed point is provided, the spatial support $R_s$ is the only hyperparameter that requires configuration. It is crucial to emphasize that $R_s$ should not be strictly equated to the target's physical equivalent radius $r$. 

The role of $R_s$ is to determine how early the geometric penalty dominates the optimization. If $R_s$ is too small, the term $-d_{max}^2/(2R_s^2)$ increases too rapidly and causes premature stopping, leading to under-grown masks. If $R_s$ is too large, the prior becomes overly weak and background leakage increases. Therefore, $R_s$ should be interpreted as a tolerance scale of the geometric prior rather than as the physical target radius itself. Since infrared small targets usually exhibit blurred transition regions beyond the compact core, the admissible support is generally larger than $r$. When a scale estimate is available, it is natural to let $R_s$ increase with $r$; in practice, we use the form $R_s = k \cdot r$ with $k>1$. The exact value of $k$ does not need to be precise, provided that the geometric prior remains within a reasonable operating range.

This characteristic endows our framework with high flexibility across different application phases:
\begin{itemize}
    \item Annotation Phase (Blind Mode): Since the instance-specific radius is unavailable during manual point-labeling, we adopt a robust default $R_s$ based on the typical size range of infrared small targets.
    \item Inference Phase (Guided Mode): Once the target scale $\hat{r}$ is predicted by RPR-Net, $R_s$ can be adaptively determined to improve mask recovery.
\end{itemize}

From this perspective, initialization and spatial support play complementary roles: the seed point affects the early growth trajectory, whereas $R_s$ determines the admissible spatial range of that growth. The remaining question is whether the resulting energy evolution admits a finite optimum rather than diverging indefinitely.

\paragraph{Analytical Conditions for Energy Convergence}

PAMG does not rely on an online stopping criterion.
Instead, the energy value is recorded at each expansion step and the optimal
mask is selected a posteriori as the maximal configuration along the
entire growth trajectory.
Therefore, the theoretical analysis focuses on the existence of a finite
energy maximizer on the discrete sequence
$\{\mathcal{E}(S_n)\}$.

Based on a discrete energy increment analysis (detailed in Appendix~\ref{app:energy_analysis}),
the energy is guaranteed to attain a finite maximum under mild conditions.
Specifically, the following detectability boundary is obtained:
\begin{equation}
\mathrm{SCR}
>
\frac{1}{\gamma}
\sqrt{
2n
\max
\left(
0,
\frac{1}{n \ln n}
-
\frac{1}{2\pi R_s^2}
\right)
}.
\label{equ:generation_boundary}
\end{equation}
where $\mathrm{SCR} = |\mu_T - \mu_B| / \sigma_B$ denotes the signal-to-clutter ratio, and $\gamma = \sigma_B / \sigma_T$ denotes the uniformity ratio, quantifying the relative smoothness between the background and the target.

This inequality implies that the lower bound of the SCR required for effective mask generation is jointly determined by the textural uniformity and the target size. Specifically, as the target size $n$ increases and the uniformity contrast becomes more pronounced (i.e., a larger $\gamma$, representing a smoother target amidst fluctuating clutter), the required SCR threshold decreases. Conversely, when the target size is small or the target texture resembles the background ($\gamma \approx 1$), this uniformity gain diminishes, thereby necessitating a higher SCR for reliable generation.

\subsection{Radius-aware Point Regression Network (RPR-Net)}
\label{subsec:network}

To achieve efficient localization and scale perception for infrared small targets, we propose the RPR-Net. As illustrated in Fig. \ref{fig:overview}, RPR-Net reformulates detection as an end-to-end spatiotemporal feature extraction and geometric parameter regression process.

\subsubsection{Input Strategy and Siamese Feature Extraction}
Infrared small targets often appear as textureless blobs in single frames, easily submerged by complex backgrounds. While data exists as video streams, efficiently exploiting temporal cues remains a challenge. We adopt a bilateral temporal context strategy, constructing an input triplet $\{I_{t-1}, I_t, I_{t+1}\}$ with the current frame $I_t$ as the keyframe. Compared to methods using long sequences \cite{chen2024sstnet,zhu2024tmp,duan2024tripledomain}, this strategy minimizes computational overhead and latency while utilizing the future frame $I_{t+1}$ to capture instantaneous motion trends for bidirectional verification.

To prevent the premature coupling of target and background information caused by early channel stacking, we design a Siamese Backbone with shared weights. The triplet frames are processed in parallel to extract independent spatial features $f_{t-1}, f_t, f_{t+1}$. Regarding the backbone architecture, we employ a trimmed CSPDarknet-Tiny. Considering that 32$\times$ downsampling results in the physical loss of small target features and introduces excessive receptive field noise during upsampling, we remove the deep stage (Stage 5) of the backbone, restricting the maximum downsampling rate to 16$\times$. Subsequently, a Feature Pyramid Network (FPN) is introduced to fuse multi-scale features, finally outputting high-resolution feature maps at 4$\times$ downsampling, thereby preserving precise spatial localization details.

\subsubsection{Temporal Difference Attention}
To capture motion cues without heavy computation, we propose the Temporal Difference Attention (TDA) module. This module enhances the current frame feature $f_t$ by highlighting regions with significant motion relative to neighbors.
First, we compute the bilateral feature difference:
\begin{equation}
    D_t = \mathcal{C}_{mot}(|f_t - f_{t-1}| + |f_t - f_{t+1}|)
\end{equation}
where $\mathcal{C}_{mot}$ denotes a motion extractor composed of a $1 \times 1$ convolution and a Sigmoid activation. 

To suppress false motion responses caused by camera jitter or abrupt view changes, we introduce a learnable global gating mechanism. Specifically, we measure the scene instability by calculating the global jitter intensity $\mathcal{J}$. This is formulated as the Global Average Pooling (GAP) of the absolute difference between the future frame $f_{t+1}$ and the past frame $f_{t-1}$:

\begin{equation}
    \mathcal{J} = \text{GAP}(|f_{t+1} - f_{t-1}|)
\end{equation}

Based on $\mathcal{J}$, we dynamically adjust the attention weight using a learnable temperature coefficient $\alpha$:
\begin{equation}
    G = \exp(-\alpha \cdot \mathcal{J}), \quad f_{t}^{'} = f_t + G \cdot D_t \cdot f_t
\end{equation}
Here, $f_{t}^{'}$ represents the motion-enhanced feature of the current frame. When the scene is stable, the network utilizes motion cues; when severe jitter occurs, $G \rightarrow 0$, and the module adaptively degrades to rely on spatial appearance.

After enhancing the current frame, we design a Temporal Feature Fusion module to synthesize the temporal context $\{f_{t-1}, f_{t}^{'}, f_{t+1}\}$ via content-aware re-weighting.

We first concatenate the three feature maps along the channel dimension to form $F_{stack} \in \mathbb{R}^{3C \times H \times W}$. To capture the global importance of each frame, we apply GAP followed by a Multi-Layer Perceptron (MLP) to generate a channel-wise weight vector $W$:
\begin{equation}
    W = \sigma(\mathbf{W}_2 \delta(\mathbf{W}_1(\text{GAP}(F_{stack}))))
\end{equation}
where $\delta$ is the ReLU activation, $\sigma$ is the Sigmoid function, and $\mathbf{W}_1, \mathbf{W}_2$ are learnable weights of the MLP. The generated weight vector $W$ is then split into three parts $w_{t-1}, w_{t}, w_{t+1}$ corresponding to the past, current, and future frames.

The final fused spatiotemporal feature $F_{temp}$ is obtained by a weighted summation followed by a smoothing convolution:
\begin{equation}
    F_{temp} = \text{Conv}_{3\times3}(w_{t-1} \cdot f_{t-1} + w_{t} \cdot f_{t}^{'} + w_{t+1} \cdot f_{t+1})
\end{equation}
This mechanism allows the network to adaptively emphasize the most informative frames before spatial processing.

\subsubsection{Spatial Attention Block}
After temporal fusion, residual background clutter may still exist. To further suppress noise and highlight the target's informative regions, we apply a standard spatial attention block (SAB). The SAB generates a spatial weight map $M_s$ via pooling and convolution operations:
\begin{equation}
    M_s = \sigma(\mathcal{C}_{7\times7}([\text{AvgPool}(F_{temp}); \text{MaxPool}(F_{temp})]))
\end{equation}
The final feature map for detection is obtained by $F_{final} = F_{temp} \cdot M_s$.

\subsubsection{Decoupled Heads and Adaptive Label Assignment}
We employ three decoupled lightweight heads to predict the target geometry from $F_{final}$. Instead of regressing bounding box dimensions, we simplify the task to predicting the "center point + effective radius":
\begin{itemize}
    \item Heatmap Head: Predicts a probability map of size $H/4 \times W/4$ to localize target centers.
    \item Offset Head: Regresses $(\delta x, \delta y)$ to compensate for quantization errors.
    \item Radius Head: Directly predicts the scalar effective radius $r$.
\end{itemize}

During the training phase, we implement an adaptive Gaussian label assignment strategy. For each ground-truth target, the standard deviation $\sigma$ of the Gaussian kernel is dynamically adjusted based on its physical radius $r_{gt}$:
\begin{equation}
    \sigma = \max(\sigma_{min}, \min(r_{gt}, \sigma_{max}))
\end{equation}
where $\sigma_{min}$ and $\sigma_{max}$ denote the lower and upper bounds of the standard deviation, respectively. This mechanism ensures that the network receives sufficient positive supervision for extremely small targets while accurately modeling the spatial energy distribution of larger ones. For heatmaps under a $4\times$ downsampling ratio, we empirically set $\sigma_{min} = 2$ and $\sigma_{max} = 4$.

 \subsubsection{Loss Function}
To train the RPR-Net end-to-end, we employ a multi-task loss function defined as:
\begin{equation}
    L_{total} = \lambda_{hm} L_{heat} + \lambda_{off} L_{off} + \lambda_{rad} L_{rad}
\label{equ:loss_func}
\end{equation}
where $L_{heat}$ is the focal loss utilized to handle the sample imbalance between sparse targets and dense backgrounds. $L_{off}$ and $L_{rad}$ represent the $L_1$ regression losses for the quantization offset and scalar radius, respectively. Note that the regression losses are computed exclusively at the ground truth centroids (i.e., Masked $L_1$ Loss).

\subsection{Synergistic Training and Inference}
\label{subsec:synergy}

The proposed framework establishes a robust closed-loop mechanism between data generation and network prediction, operating in two distinct phases:

\subsubsection{Training Phase: Mask-to-Point Supervision}
In this phase, the data flow is from PAMG to RPR-Net. Since manual point annotations lack geometric information, we utilize PAMG (Section \ref{subsec:PAMG}) to offline generate high-fidelity pseudo-masks for the training set. From these masks, we extract the precise centroid coordinates $(x_{gt}, y_{gt})$ and the effective physical radius $r_{gt} = \sqrt{Area/\pi}$. These geometric attributes serve as ground-truth supervision for training RPR-Net via the loss function defined in Eq.\ref{equ:loss_func}. This strategy effectively transfers the physics-informed shape knowledge from the PAMG algorithm into the lightweight detection network.

\subsubsection{Inference Phase: Point-to-Mask Detection}
In this phase, the data flow reverses from RPR-Net to PAMG. RPR-Net efficiently processes the image sequence and outputs the target triplet $(\hat{x}, \hat{y}, \hat{r})$. For standard detection tasks, these parameters provide sufficient localization and scale estimation. However, for tasks requiring fine-grained pixel-level perception, we employ the predicted triplet to initialize the PAMG algorithm. Specifically, the predicted effective radius $\hat{r}$ is used as a scale cue to construct the PAMG spatial support parameter $R_s$. In practice, we use a proportional mapping $R_s=k\cdot\hat{r}$. This allows the system to recover fine-grained target shapes solely from the network's regression outputs, achieving "segmentation-level" performance without the computational burden of a heavy segmentation model.

\subsection{Construction of SIRSTD-Pixel Dataset}
\label{sec:dataset_construction}

To support pixel-level evaluation in sequential infrared small target detection, we adopt the SIRSTD dataset \cite{tong2024sttrans} as the base source for further annotation. Originally collected during the Anti-UAV competition \cite{jiang2023antiuav}, this dataset contains real-world long-wave infrared (LWIR) sequences with a resolution of $640 \times 512$ pixels and covers diverse scenes such as urban, forest, maritime, and sky backgrounds.

Although the raw dataset provides rich temporal information, it is originally annotated only with bounding boxes. Since box annotations are insufficient for evaluating pseudo-mask quality, compact support recovery, and geometry-aware supervision, we further annotate the dataset at the pixel level and construct SIRSTD-Pixel. The annotation pipeline and the resulting dataset statistics are described below.

\subsubsection{Semi-automated Annotation Pipeline}

Given a dataset size of more than 50,000 frames, purely manual pixel-wise annotation would be time-consuming and difficult to keep consistent. To improve efficiency while maintaining annotation quality, we design a semi-automated human-in-the-loop pipeline. The workflow consists of three phases.

\paragraph{Point-supervised Candidate Mask Generation}
We leverage the existing bounding boxes in the raw dataset as coarse priors to reduce the burden of manual target localization. Specifically, we use the geometric center of each box as the initial seed point and employ PAMG (Algorithm \ref{alg:PAMG}) to generate a candidate mask. This step provides a pre-annotation result for subsequent human verification and refinement.

\paragraph{Development of Custom Tool: Label-IRST}
General-purpose annotation tools such as LabelMe and CVAT are mainly designed for polygon-based annotation of macroscopic objects. They are less suitable for infrared small targets, which occupy only a few pixels and often have ambiguous boundaries. In addition, annotation on raw infrared imagery usually requires frequent zooming and repeated local inspection, while the built-in automation functions in existing tools are not tailored to this task.

\begin{figure}[t]
    \centering
    \includegraphics[width=1\linewidth]{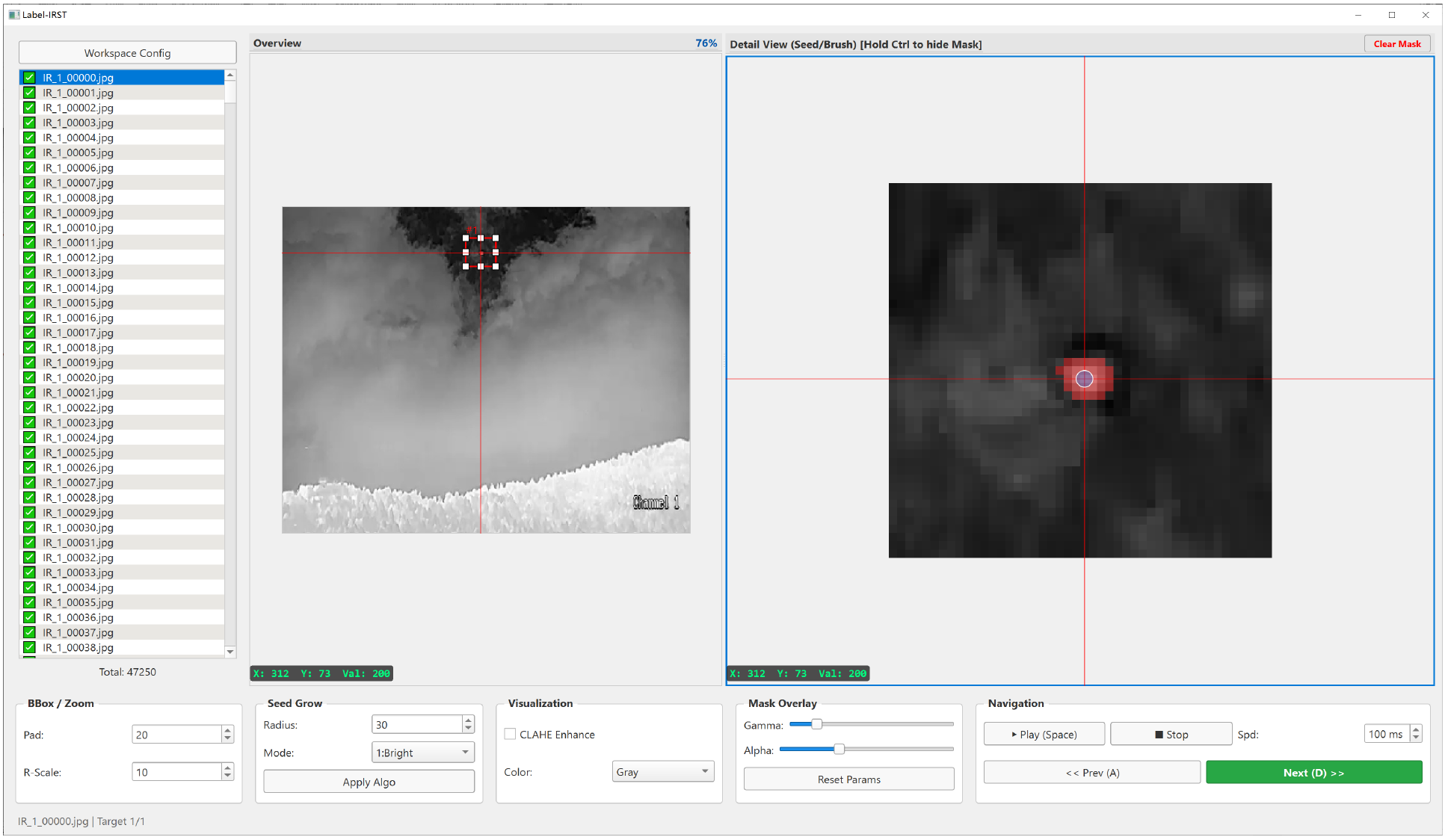}
    \caption{The graphical user interface of the proposed Label-IRST annotation tool. The interface features a dual-view mechanism designed to reduce visual fatigue and unnecessary cursor movement. The Global View (left) displays the full-resolution raw infrared image for rapid Region of Interest selection. The Detail View (right) automatically renders the selected area in high magnification, allowing users to verify and refine the candidate mask (shown in red). The bottom panel provides controls for annotation parameters and visual enhancement settings.}
    \label{fig:label_irst}
\end{figure}

To address these limitations, we developed a specialized interactive annotation tool named Label-IRST, as illustrated in Fig. \ref{fig:label_irst}. The tool includes several functions tailored to infrared small target annotation:
\begin{itemize}
    \item Automated mask initialization: Label-IRST integrates PAMG so that a candidate mask can be generated from a selected seed point and then refined when necessary.
    \item Dual-view interaction: The left panel displays the full image for region selection, while the right panel shows a magnified local view for fine annotation.
    \item Visual enhancement: Local contrast enhancement and pseudo-color rendering are provided to improve the visibility of weak targets and surrounding boundaries.
\end{itemize}
Label-IRST has been open-sourced for research use, and the implementation details are available in our GitHub repository.

\paragraph{Manual Verification and Refinement}
With the aid of Label-IRST, annotators mainly act as reviewers rather than drawing masks from scratch. For accurately generated masks, they can directly confirm the result; for inaccurate or missed targets, they perform local refinement. Compared with fully manual annotation, this semi-automated workflow reduces annotation effort and also helps improve consistency across a large dataset.

\subsubsection{Statistical Analysis of SIRSTD-Pixel Dataset}
\begin{figure}[t]
    \centering
    \includegraphics[width=1\linewidth]{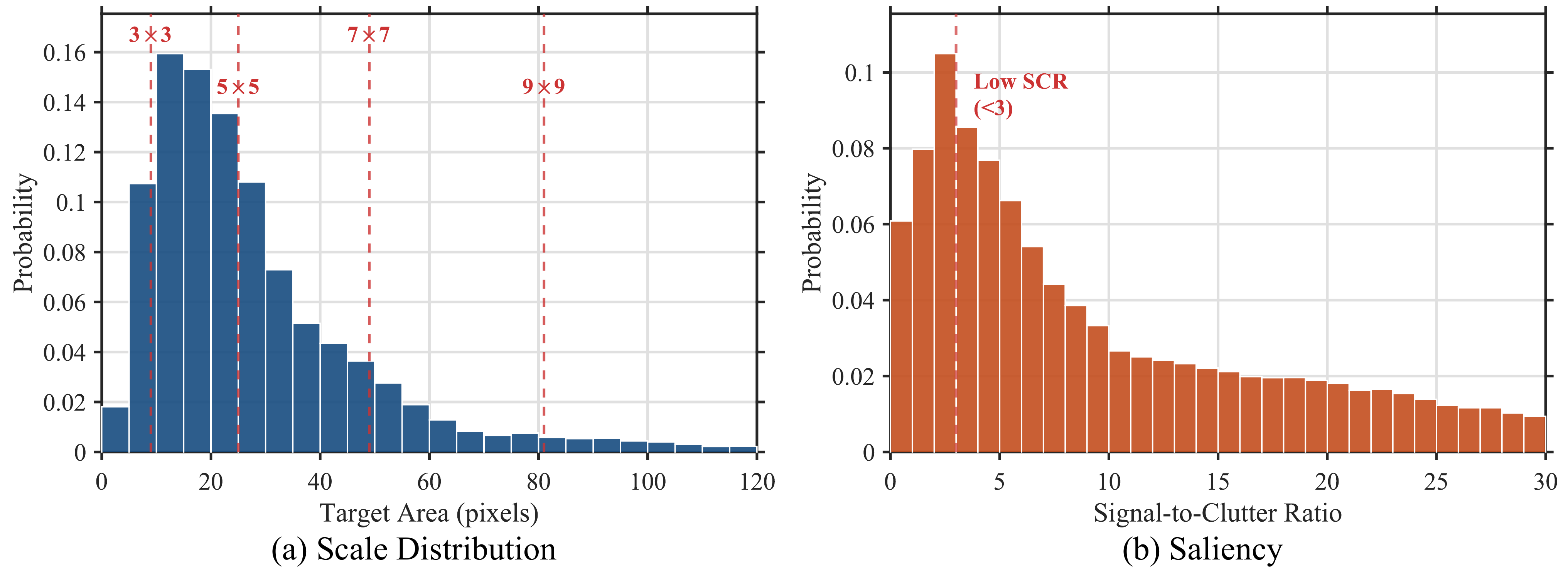} 
    \caption{Statistical characteristics of the SIRSTD-Pixel dataset. (a) Scale distribution: target areas are mainly concentrated within a small pixel range, consistent with infrared small target characteristics. (b) Saliency distribution: the SCR values span a broad range, reflecting the diversity of target saliency in the dataset.}
    \label{fig:sirstd_stats}
\end{figure}

Upon completing the pixel-level annotation, we established the SIRSTD-Pixel dataset comprising 47,250 frames. Based on the annotated masks, we summarize two basic statistical properties of the targets, namely scale and saliency, as shown in Fig. \ref{fig:sirstd_stats}.

\begin{itemize}
    \item Scale distribution: As shown in Fig. \ref{fig:sirstd_stats}(a), the target area is mainly concentrated in a small pixel range, with the probability density peaking around $3 \times 3$ to $5 \times 5$ pixels. Most samples fall below $9 \times 9$ pixels, which is consistent with the definition of infrared small targets.
    \item Saliency distribution: As shown in Fig. \ref{fig:sirstd_stats}(b), the SCR values span a broad range, indicating that the dataset contains targets with different saliency levels. This distribution helps characterize the diversity of target visibility in realistic infrared scenes.
\end{itemize}

\section{Experimental Analysis}
\label{sec:experiments}

\subsection{Experimental Setup}

In this section, we conduct comprehensive experiments to evaluate the proposed Point-to-Mask framework. We first examine the quality of PAMG-generated pseudo masks to verify whether arbitrary point annotations can provide reliable geometric supervision. We then compare the proposed framework with representative state-of-the-art methods under both fully supervised and point-supervised settings through quantitative and qualitative analyses. 
Further experiments investigate the robustness and interpretability of PAMG via sensitivity analysis, component ablation, and empirical validation of the derived generation boundary. We also perform ablation studies on RPR-Net to analyze the contributions of temporal modeling and backbone design. Finally, we evaluate the efficiency of the overall framework in terms of model complexity and inference latency.

\subsubsection{Datasets}
To evaluate the detection performance under different scenarios, we conduct experiments on one large-scale sequential dataset and two widely used single-frame datasets. The constructed SIRSTD-Pixel serves as the primary benchmark. Compared with existing datasets, it provides dense pixel-level annotations for continuous infrared sequences, which enables the evaluation of spatiotemporal algorithms. The dataset is divided into a training set of 22 sequences (26,784 frames) and a testing set of 17 sequences (20,466 frames).

To further examine the generalization ability of the proposed framework, we additionally conduct experiments on two large-scale single-frame datasets, NUDT-SIRST \cite{li2022dense} and IRSTD-1K \cite{zhang2022isnet}. NUDT-SIRST contains 1,327 images with diverse backgrounds and targets of multiple scales, while IRSTD-1K includes 1,001 images with various clutter types. For both datasets, we follow the commonly used training and validation splits reported in \cite{li2022dense,zhang2022isnet,li2023monte} to ensure fair comparison.

\subsubsection{Evaluation Metrics}
We adopt the standard metrics in infrared small target detection, including Probability of Detection ($P_d$), False Alarm Rate ($F_a$), Area Under Curve (AUC), and Intersection over Union (IoU).

$P_d$ (also known as Recall) is a target-level metric measuring the capability to correctly locate targets. $F_a$ is a pixel-level metric measuring the suppression of false alarms. They are defined as:
\begin{equation}
    P_d = \frac{N_{TP}}{N_{total}}, \quad F_a = \frac{N_{FP}}{N_{pixels}}
    \label{eq:pd_fa}
\end{equation}
where $N_{TP}$ is the number of correctly predicted targets, and $N_{total}$ is the total number of real targets. $N_{FP}$ denotes the number of background pixels incorrectly predicted as targets, and $N_{pixels}$ represents the number of actual background pixels.
The receiver operating characteristic (ROC) curve plots $P_d$ against $F_a$, and AUC represents the area under this curve. A larger AUC value indicates better overall detection performance.

To evaluate the precision of the recovered masks, we use the IoU, defined as:
\begin{equation}
    \text{IoU} = \frac{TP}{TP + FP + FN}
\end{equation}
where $TP, FP,$ and $FN$ denote True Positives, False Positives, and False Negatives at the pixel level, respectively. Mean IoU (mIoU) is calculated as the average IoU over all test samples.

\subsubsection{Implementation Details}
All deep learning-based methods in this paper were implemented using the PyTorch framework and executed on a single NVIDIA GeForce RTX 4090 GPU. The input to the proposed RPR-Net consists of three consecutive infrared frames, each resized to $512 \times 512$ pixels. During training, we employ the AdamW optimizer with an initial learning rate of $1 \times 10^{-3}$ and a weight decay of $1 \times 10^{-4}$ to ensure stable convergence. The learning rate is adjusted by a cosine annealing scheduler and decays to a minimum of $1 \times 10^{-5}$ over 100 epochs. Regarding the hyper-parameters in the loss function, the weighting factors are empirically set to $\lambda_{hm}=1.0$, $\lambda_{off}=1.0$, and $\lambda_{rad}=0.1$ to balance the gradient magnitudes between heatmap classification and geometric regression. 
For PAMG under the guided mode, where the target radius $\hat{r}$ is provided by RPR-Net, we set the spatial support parameter as $R_s = k \cdot \hat{r}$ with $k=5$. 
For a fair comparison, all baseline methods are retrained using their official open-source implementations with default configurations.

\subsection{Evaluation of Pseudo-Label Generation Quality}

In point-supervised infrared small target detection frameworks, the quality of pseudo-labels generated from single-point annotations determines the upper bound of detection performance. Therefore, in this section, we quantitatively evaluate the generation quality of our method against mainstream training-free mask generation techniques (MCLC\cite{li2023monte}, MCGC\cite{kou2024mcgc}, HMG\cite{he2025hybrid}) and the Segment Anything Model (SAM\cite{kirillov2023segment}, ViT-H version) across the SIRSTD, NUDT-SIRST, and IRSTD-1K datasets.

\subsubsection{Evaluation Protocol and Settings}
To quantify the overlap between the generated masks and the ground truth (GT), we employ the IoU metric. 

Although GT serves as the benchmark, manual annotation inevitably introduces subjective bias \cite{kou2024mcgc}. For tiny targets (e.g., $<3 \times 3$ pixels), a deviation of merely 1-2 pixels can cause drastic fluctuations in IoU. Nevertheless, this remains the most direct and objective quantitative assessment available.

To rigorously assess the potential of algorithms in real-world point supervision, we designed two evaluation modes:
\begin{itemize}
    \item Prior Mode: Algorithms are permitted to utilize built-in prior information (e.g., target size priors for MCLC/MCGC, optimal seed points for HMG/SAM). Parameter settings follow the official implementations of the respective methods.
    \item Blind Mode: This mode strictly simulates a realistic point-supervision scenario where only a single random coordinate is provided, devoid of any size or shape priors.
\end{itemize}

In Blind mode, MCLC employs a fixed $32 \times 32$ crop window; MCGC generates a simulation box with a random margin of $3 \sim 15$ pixels around the target; HMG and SAM rely solely on random point prompts. In contrast, our PAMG method requires no size or location priors, utilizing a fixed $R_s=20$ and random seed points. This value is not engineered for a specific dataset; rather, it is an empirical choice based on the typical size
range of infrared small targets. The sensitivity of this parameter is further investigated in the ablation studies in Sec. \ref{subsec:pamg_ablation}. To ensure reproducibility, the random seed is fixed at 3407.

\begin{table}[t]
  \centering
  \caption{Quantitative comparison of pseudo-label generation quality on the SIRSTD, NUDT-SIRST, and IRSTD-1K datasets. All values are reported in mIoU (\%). $\Delta$ denotes the performance gap between Blind and Prior modes.}
  \label{tab:pseudo_label_quality}
  \renewcommand{\arraystretch}{1.15}
  \setlength{\tabcolsep}{4pt}
  \resizebox{\linewidth}{!}{
  \begin{tabular}{l c c c c c}
    \toprule
    \textbf{Method} & \textbf{Setting} & \textbf{SIRSTD} & \textbf{NUDT-SIRST} & \textbf{IRSTD-1K} & \textbf{Mean} \\
    \midrule
    \multirow{3}{*}{MCLC \cite{li2023monte}} 
       & Prior    & 63.08 & 69.44 & 71.33 & 67.95 \\
       & Blind    & 54.42 & 62.69 & 66.97 & 61.36 \\
       & $\Delta$ & -8.66 & -6.75 & -4.36 & -6.59 \\
    \midrule
    \multirow{3}{*}{MCGC \cite{kou2024mcgc}} 
       & Prior    & 63.58 & 73.04 & 63.62 & 66.75 \\
       & Blind    & 61.01 & 70.93 & 63.26 & 65.07 \\
       & $\Delta$ & -2.57 & -2.11 & -0.36 & -1.68 \\
    \midrule
    \multirow{3}{*}{HMG \cite{he2025hybrid}} 
       & Prior    & 51.11 & 69.19 & 61.75 & 60.68 \\
       & Blind    & 43.04 & 55.95 & 50.91 & 49.97 \\
       & $\Delta$ & -8.07 & -13.24 & -10.84 & -10.71 \\
    \midrule
    \multirow{3}{*}{SAM \cite{kirillov2023segment}} 
       & Prior    & 34.38 & 66.28 & 57.53 & 52.73 \\
       & Blind    & 33.63 & 66.15 & 56.84 & 52.21 \\
       & $\Delta$ & -0.75 & -0.13 & -0.69 & -0.52 \\
    \midrule
    PAMG 
       & Blind    & \textbf{88.34} & \textbf{75.11} & \textbf{71.63} & \textbf{78.36} \\
    \bottomrule
  \end{tabular}
  }
\end{table}

\subsubsection{Analysis of Results}

Table \ref{tab:pseudo_label_quality} reports the quantitative comparison of different training-free pseudo-label generation methods across three datasets. As shown, PAMG achieves the highest average mIoU among the compared methods, reaching 78.36\%, higher than the prior-dependent methods MCLC (67.95\%) and MCGC (66.75\%). This result suggests that PAMG captures several relevant characteristics of infrared small targets. 

Further insights can be obtained by examining the performance differences between the Prior and Blind modes. Existing methods generally show varying degrees of dependence on additional prior information. For example, the performance of MCLC drops by 6.59\% in the Blind setting, suggesting that a fixed cropping window is difficult to accommodate targets with different scales. HMG exhibits an even larger degradation of 10.71\%, indicating strong sensitivity to the initialization of the seed point. Although SAM demonstrates strong general segmentation capability, its average mIoU on infrared small target datasets is only 52.73\%, implying that general vision foundation models cannot directly bridge the imaging differences between this domain and natural image segmentation. In contrast, PAMG consistently achieves the best performance without relying on any size or shape priors, suggesting that the modeling of local saliency, internal homogeneity, and spatial compactness in its energy formulation is effective.

PAMG achieves an mIoU of 88.34\% on the SIRSTD-Pixel dataset, which is noticeably higher than its performance on NUDT-SIRST and IRSTD-1K. This difference is partly related to the annotation process of SIRSTD-Pixel, where a Human-in-the-Loop strategy was adopted during dataset construction, resulting in stronger consistency between human annotations and the regions recovered by PAMG. The result also indicates that the regions recovered by PAMG largely coincide with the targets identified by human annotators based on infrared responses. These observations support the use of PAMG as a pseudo-label generator and suggest that it may be useful for assisting large-scale pixel-level annotation of infrared datasets.

\subsection{Comparison with State-of-the-Art Methods}

To comprehensively evaluate the effectiveness of the proposed framework, we conduct systematic comparisons with several representative infrared small target detection methods. Specifically, under the fully supervised setting, pixel-level GT is used to extract target geometric information (center points and equivalent radii) for training RPR-Net, allowing us to assess the upper-bound performance of the network architecture itself. Under the point-supervised setting, only a single randomly annotated point is provided for each target, while PAMG is employed to generate pseudo masks and the corresponding geometric supervision signals, thereby evaluating the practical feasibility of the framework under low-cost annotation conditions.

The compared methods include mainstream single-frame pixel-level modeling approaches, such as ACM \cite{dai2021asymmetric}, AGPCNet \cite{zhang2023attentionguided}, ALCNet \cite{dai2021attentional}, DNANet \cite{li2022dense}, ISNet \cite{zhang2022isnet}, ISTDU-Net \cite{hou2022istdunet}, MSHNet \cite{liu2024infrared}, RDIAN \cite{sun2023receptivefield}, and SCTransNet \cite{yuan2024sctransnet}, as well as temporal detection methods including SSTNet \cite{chen2024sstnet}, DTUM \cite{li2025directioncoded}, and TMP \cite{zhu2024tmp}. SIRSTD-Pixel serves as the primary benchmark in this work. This dataset contains complex backgrounds, real continuous sequences, and pixel-level annotations, enabling a comprehensive evaluation of the core idea of the proposed framework. In contrast, experiments on IRSTD-1K and NUDT-SIRST are mainly conducted to examine the generalization capability of the framework in static single-frame scenarios.
\subsubsection{Quantitative Analysis}

\begin{table*}[t]
\centering
\caption{Quantitative comparison with state-of-the-art methods. The best results are highlighted in \textbf{bold}. Methods prefixed with \textit{Ps-} denote point-supervised approaches. mIoU and $P_d$ are reported in (\%), and $F_a$ is reported in $10^{-5}$.}
\label{table:MainResults}

\renewcommand{\arraystretch}{1.15}
\setlength{\tabcolsep}{4pt}

\resizebox{\textwidth}{!}{
\begin{tabular}{cccccccccccc}
\toprule
\multirow{2}{*}{\textbf{Type}} & \multirow{2}{*}{\textbf{Method}} & \multirow{2}{*}{\textbf{Sup.}} & \multirow{2}{*}{\textbf{Out}} 
& \multicolumn{4}{c}{\textbf{SIRSTD-Pixel}} 
& \multicolumn{4}{c}{\textbf{IRSTD-1K / NUDT-SIRST}} \\

\cmidrule(lr){5-8}\cmidrule(lr){9-12}

& & & 
& \textbf{mIoU$\uparrow$} & \textbf{$P_d\uparrow$} & \textbf{$F_a\downarrow$} & \textbf{AUC$\uparrow$}
& \textbf{mIoU} & \textbf{$P_d$} & \textbf{$F_a$} & \textbf{AUC} \\

\midrule

\multirow{9}{*}{\rotatebox{90}{Single-frame}}
& ACM \cite{dai2021asymmetric} & Pixel & Mask 
& 43.13 & 69.04 & 1.80 & 0.8452 
& 58.34 / 65.65 & 94.59 / 97.56 & 5.23 / 3.05 & 0.9730 / 0.9878 \\

& AGPCNet \cite{zhang2023attentionguided} & Pixel & Mask 
& 54.55 & 71.42 & 0.33 & 0.8571 
& 64.66 / 79.78 & 96.28 / 99.15 & 3.22 / 2.01 & 0.9814 / 0.9958 \\

& ALCNet \cite{dai2021attentional} & Pixel & Mask 
& 45.16 & 75.50 & 3.11 & 0.8775 
& 59.61 / 52.85 & 96.62 / 98.20 & 5.12 / 8.57 & 0.9831 / 0.9910 \\

& DNANet \cite{li2022dense} & Pixel & Mask 
& 55.06 & 67.81 & \textbf{0.26} & 0.8390 
& 68.92 / 84.50 & 94.26 / 98.73 & 2.58 / 0.82 & 0.9713 / 0.9936 \\

& ISNet \cite{zhang2022isnet} & Pixel & Mask 
& 47.14 & 74.19 & 3.89 & 0.8710 
& 64.01 / 61.57 & 95.27 / 98.52 & 3.74 / 20.5 & 0.9763 / 0.9926 \\

& ISTDU-Net \cite{hou2022istdunet} & Pixel & Mask 
& 53.43 & 71.53 & 0.88 & 0.8577 
& 63.45 / 83.32 & 96.62 / 98.20 & 6.54 / 1.00 & 0.9831 / 0.9910 \\

& MSHNet \cite{liu2024infrared} & Pixel & Mask 
& 48.08 & 69.28 & 5.21 & 0.8464 
& 69.68 / 82.85 & 93.92 / 98.62 & 1.28 / 1.53 & 0.9696 / 0.9931 \\

& RDIAN \cite{sun2023receptivefield} & Pixel & Mask 
& 49.20 & 72.89 & 3.07 & 0.8645 
& 65.06 / 82.06 & 90.88 / 98.83 & 2.48 / 2.24 & 0.9544 / 0.9942 \\

& SCTransNet \cite{yuan2024sctransnet} & Pixel & Mask 
& 48.81 & 72.52 & 3.92 & 0.8626 
& 66.58 / 83.53 & 96.62 / 98.73 & 3.79 / 1.42 & 0.9831 / 0.9936 \\

\midrule

\multirow{7}{*}{\rotatebox{90}{Sequential}}
& DTUM \cite{li2025directioncoded} & Pixel & Mask 
& 55.99 & 75.72 & 0.72 & 0.8786 
& \multicolumn{4}{c}{N/A} \\

& SSTNet \cite{chen2024sstnet} & Box & Box 
& 29.99 & 83.33 & 18.2 & 0.9166 
& \multicolumn{4}{c}{N/A} \\

& TMP \cite{zhu2024tmp} & Box & Box 
& 30.72 & 85.21 & 10.9 & 0.9260 
& \multicolumn{4}{c}{N/A} \\

\cmidrule(lr){2-12}

& RPR-Net & Pixel & Circle 
& 35.44 & \textbf{91.54} & 5.41 & \textbf{0.9577} 
& 42.74 / 48.34 & 95.61 / 94.60 & 9.10 / 16.2 & 0.9780 / 0.9730 \\

& RPR-Net+PAMG & Pixel & Mask 
& \textbf{61.93} & 90.01 & 6.53 & 0.9500 
& 55.60 / 67.28 & 95.27 / 94.28 & 8.89 / 18.2 & 0.9763 / 0.9712 \\

\cmidrule(lr){2-12}

& \textit{Ps-RPR-Net} & Point & Circle 
& 32.28 & 86.68 & 7.22 & 0.9334 
& 38.73 / 47.44 & 93.92 / 92.58 & 3.83 / 5.87 & 0.9696 / 0.9629 \\

& \textit{Ps-RPR-Net+PAMG} & Point & Mask 
& 57.43 & 85.76 & 6.60 & 0.9288 
& 58.92 / 66.45 & 93.58 / 92.06 & 4.53 / 9.87 & 0.9679 / 0.9602 \\

\bottomrule
\end{tabular}
}
\end{table*}

Table \ref{table:MainResults} reports the overall comparison with existing methods. On the primary benchmark SIRSTD-Pixel, RPR-Net achieves the highest target-level detection performance, with a $P_d$ of 91.54\% and an AUC of 0.9577. This result indicates that the proposed temporal modeling and geometric regression scheme is effective for sequential infrared small target detection.

PAMG significantly improves the mask-level expressiveness of the geometric predictions. Under full supervision, RPR-Net+PAMG achieves the best mIoU of 61.93\% on SIRSTD-Pixel while maintaining competitive detection performance ($P_d=90.01\%$, AUC $=0.9500$). The circle predicted by RPR-Net provides only a compact geometric descriptor rather than a dense shape representation, which limits its mIoU when evaluated directly. By recovering the target support from the predicted center and radius, PAMG substantially improves the mask quality, indicating that the geometric predictions remain sufficiently informative for mask reconstruction.
This comparison also highlights the advantage of the circle-based representation for IRSTD. Although both circle and box outputs belong to geometric prediction, the masks derived from our circle representation are noticeably better than those from sequential box-based detectors such as SSTNet and TMP. This suggests that, for compact infrared small targets, a center–radius representation better matches the underlying target support than rectangular boxes while introducing less background redundancy.

The point-supervised results further illustrate the practical value of the framework. On SIRSTD-Pixel, \textit{Ps-RPR-Net} shows a 4.86-point drop in $P_d$ and a 0.0243 drop in AUC compared with the fully supervised RPR-Net. After incorporating PAMG, \textit{Ps-RPR-Net+PAMG} still achieves an mIoU of 57.43\%, with the gap to the fully supervised counterpart limited to 4.50 points. These results suggest that point supervision, combined with PAMG, can preserve much of the performance of full supervision at lower annotation cost.

On the single-frame datasets IRSTD-1K and NUDT-SIRST, the temporal module is disabled, and the model degenerates into a purely spatial point regression structure. In this setting, the proposed method is more oriented toward target-level localization than dense shape fitting, so a gap in mask-level accuracy compared with fully supervised segmentation models is expected. Without temporal cues, the estimation of target center and effective radius becomes more sensitive to background clutter and local shape variation. Even so, the proposed method remains competitive on target-level metrics. For example, on IRSTD-1K, RPR-Net reaches a $P_d$ of 95.61\% with an AUC of 0.9780, while remaining competitive with several fully supervised baselines. The relatively lower performance on NUDT-SIRST is likely related to its synthetic target generation process and limited training scale, which together make the spatial appearance statistics less consistent with those in the testing set and increase the difficulty of compact geometric modeling.

The results in Table \ref{table:MainResults} indicate that the behavior of the proposed method depends on the complete pipeline: temporal modeling improves target-level detection, PAMG enhances shape recovery, and point supervision remains reasonably close to full supervision. The framework performs well on the sequential benchmark while maintaining acceptable generalization in static single-frame scenarios.

\subsubsection{Qualitative Visualization Analysis}

\begin{figure*}[t]
    \centering
    \includegraphics[width=1\linewidth]{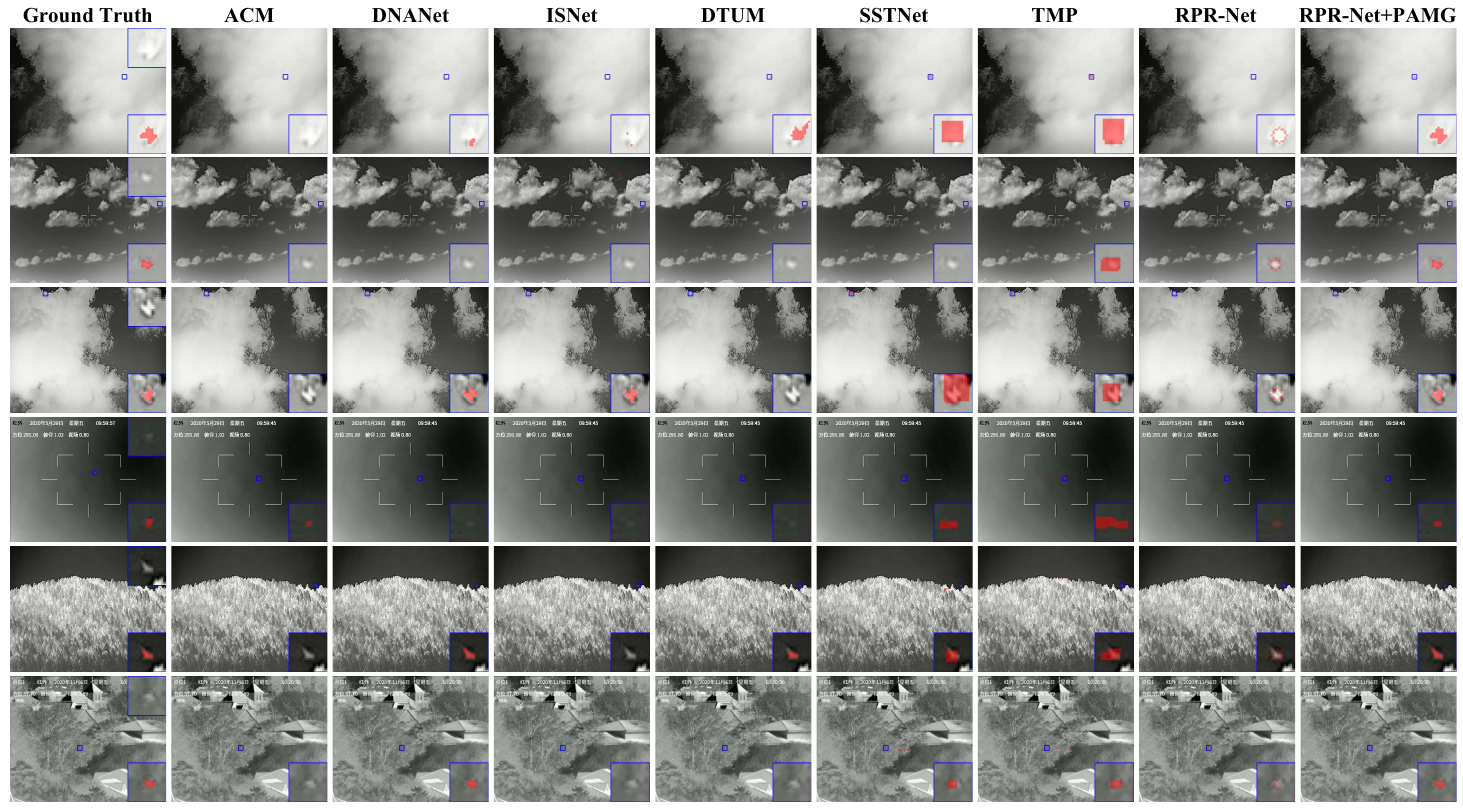}
    \caption{Qualitative comparison of representative detection results on the SIRSTD-Pixel dataset. The GT panel shows the target location (top-right) and the corresponding ground-truth mask (bottom-right), displayed as a semi-transparent red overlay. Detection results from different methods are also visualized using semi-transparent red masks. Existing methods mainly suffer from three typical failure modes: missed detections under weak target responses, false alarms triggered by complex clutter, and inaccurate shape recovery with fragmented or biased masks. In the shown examples, the proposed method localizes targets more consistently and produces masks that more closely match the annotated thermal response region.}
    \label{fig:sirstd_vis}
\end{figure*}

Fig.~\ref{fig:sirstd_vis} presents several representative visual comparisons on the SIRSTD-Pixel dataset to illustrate how different methods behave in complex scenarios. Existing approaches mainly suffer from three issues: missed detections under low contrast or weak targets, false alarms caused by bright edges, noise speckles, or structured backgrounds, and inaccurate shape recovery with fragmented or biased masks. Consequently, even when the target is detected, the resulting IoU can remain relatively low. This problem is more pronounced for methods such as SSTNet and TMP that report results in the form of bounding boxes.

In contrast, the proposed method localizes targets more consistently and produces masks that more closely match the true response region, leading to improved detection accuracy and shape completeness. This observation agrees with the quantitative results in Table~\ref{table:MainResults}. The improvement mainly comes from two aspects: the temporal modeling module enhances the spatiotemporal saliency of weakly moving targets through inter-frame differences while suppressing background-induced pseudo responses, and the physical prior introduced by PAMG allows the recovered regions to better follow the imaging characteristics and energy diffusion patterns of infrared small targets.

\subsection{Robustness Analysis of PAMG}
\label{subsec:pamg_ablation}
To further evaluate the robustness and interpretability of PAMG, we conduct parameter sensitivity analysis and component-wise ablation studies on the SIRSTD-Pixel dataset. The sensitivity analysis focuses on two factors that directly affect mask growth, namely the seed location and the spatial support parameter $R_s$, while the ablation study examines the contribution of each term in the proposed energy function.

In addition to IoU, three geometry-aware metrics are introduced to better characterize the quality of the recovered masks. The \textit{Area Ratio} is defined as $|M_{pred}|/|M_{gt}|$, where $|M_{pred}|$ and $|M_{gt}|$ denote the areas of the predicted and ground-truth masks, respectively; values close to 1 indicate good scale consistency, while smaller or larger values correspond to under-growth and over-growth. We further adopt \textit{Centroid Error}, defined as the Euclidean distance between mask centroids, and \textit{Radius Error}, defined as the absolute difference between their equivalent radii. Compared with mIoU, these metrics are less sensitive to minor boundary fluctuations and therefore better reflect whether PAMG preserves the core geometric structure required by downstream learning.

\subsubsection{Parameter Sensitivity Analysis}
We first analyze the sensitivity of PAMG to different seed locations, since point supervision provides only a single seed within each target. Three seed types are considered: Center, Random Interior, and Boundary. To reduce incidental randomness, for both Random Interior and Boundary settings we sample three points for each target and report the average result. The results in Table~\ref{tab:pamg_seed_sensitivity} show very small performance differences, with mIoU values of 87.19\%, 87.62\%, and 87.81\%, respectively. The Radius Error remains stable at about 0.26, and the Centroid Error is also similar across different seed settings. These results suggest that the seed location has little influence on the recovery of the target core or its geometric attributes.

The main variation appears in the boundary extent of the recovered masks. Center seeds produce the most conservative masks, with an Area Ratio closest to 1, while Boundary seeds generate slightly larger masks and therefore obtain a marginally higher mIoU by covering more transition pixels; Random Interior seeds fall between the two. This observation is consistent with the analysis in Sec.~\ref{sec:pamg_method}: seed location mainly affects boundary conservativeness rather than the core geometric supervision provided by PAMG. In other words, PAMG may be position-sensitive at the boundary level but remains stable at the core level, which is desirable for point-supervised training.

\begin{table}[t]
\centering
\caption{Sensitivity analysis of PAMG under different seed locations on the SIRSTD-Pixel dataset.}
\label{tab:pamg_seed_sensitivity}
\renewcommand{\arraystretch}{1.15}
\setlength{\tabcolsep}{3pt}
\resizebox{\linewidth}{!}{
\begin{tabular}{lcccc}
\toprule
\textbf{Seed Location} & \textbf{mIoU(\%)} & \textbf{Area Ratio} & \textbf{Centroid Error} & \textbf{Radius Error} \\
\midrule
Center            & 87.19 & \textbf{1.1793} & \textbf{12.1526} & \textbf{0.2635} \\
Random Interior   & 87.62 & 1.2432           & 14.6276          & 0.2673          \\
Boundary          & \textbf{87.81} & 1.2530  & 15.0028          & 0.2654          \\
\bottomrule
\end{tabular}
}
\end{table}

\begin{table*}[t]
\centering
\caption{Sensitivity analysis of the spatial support parameter in blind and guided PAMG modes on the SIRSTD-Pixel dataset. All values are reported in mIoU (\%).}
\label{tab:sensitivity}
\renewcommand{\arraystretch}{1.15}
\begin{tabular}{ccccccccccc}
\toprule
\multicolumn{11}{c}{\textit{(a) Impact of Fixed Support Radius $R_s$ (Blind Mode)}} \\
\midrule
\textbf{Radius $R_s$}
& 2 & 4 & 8 & 12 & 16 & 20 & 24 & 28 & 32 & 40 \\
\textbf{mIoU (\%)} 
& 8.75 & 34.43 & 72.69 & 85.09 & \textbf{88.37} & 88.34 & 87.73 & 87.03 & 86.25 & 84.99 \\
\midrule
\multicolumn{11}{c}{\textit{(b) Impact of Adaptive Scale Factor $k$ (Guided Mode, $R_s = k \cdot r$)}} \\
\midrule
\textbf{Scale Factor $k$} 
& 1 & 2 & 3 & 4 & 5 & 6 & 7 & 8 & 9 & 10 \\
\textbf{mIoU (\%)} 
& 8.63 & 52.53 & 82.98 & 88.21 & 89.12 & \textbf{89.25} & 88.97 & 88.74 & 88.34 & 87.96 \\
\bottomrule
\end{tabular}
\end{table*}

We further analyze the influence of the spatial support parameter $R_s$, and the results are summarized in Table~\ref{tab:sensitivity}. In blind mode, where no instance-level scale cue is available, small $R_s$ values lead to severe under-growth. For example, the mean IoU is only 8.75\% when $R_s=2$, since the geometric penalty suppresses region growth before the effective target response is fully covered. As $R_s$ increases, the performance improves rapidly and reaches a stable plateau over $R_s=16\sim24$, where the mean IoU remains around 88\%. When $R_s$ becomes overly large, the spatial constraint weakens and performance gradually decreases due to background leakage. These results indicate that $R_s$ acts as a support scale rather than an exact target radius, and PAMG remains stable within a reasonable interval.

In guided mode, the support scale is adaptively defined as $R_s = k \cdot r$ using the radius predicted by RPR-Net. Table~\ref{tab:sensitivity}(b) shows that directly setting $R_s=r$ yields only 8.63\% mIoU, indicating that the admissible support must exceed the compact target core. As $k$ increases, the performance quickly forms a plateau over $k=4\sim8$, with the best result at $k=6$. The small difference between $k=5$ and $k=6$ suggests that precise tuning of the scaling factor is unnecessary.

The trends in blind and guided modes are largely consistent. For typical infrared small targets whose radii are only a few pixels, moderate scaling in guided mode produces a support range close to the stable interval observed for fixed $R_s$ in blind mode. This also explains why a fixed default value of $R_s=20$ works well in practice, while adaptive scaling further improves mask growth when radius cues are available.

\subsubsection{Component-wise Ablation of PAMG}
\begin{figure}[t]
    \centering
    \includegraphics[width=1\linewidth]{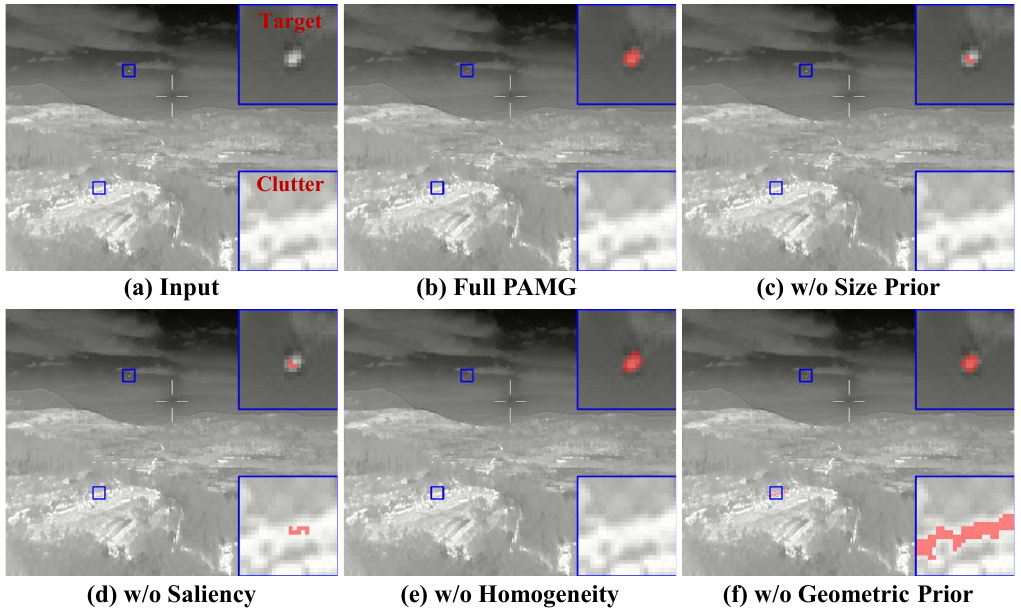}
    \caption{Visual comparison of component-wise ablations in PAMG. Red overlays denote the generated pseudo masks. (a) Input image, where the top-right inset shows a magnified view of the target and the bottom-right inset shows a magnified clutter region. (b) Full PAMG accurately recovers the compact target while suppressing the clutter region. (c) Without the size prior, the target region is under-grown. (d) Without the saliency term, the target grows incompletely while clutter expands irregularly. (e) Without the homogeneity term, the target region expands into the background clutter. (f) Without the geometric prior, the clutter region grows without effective constraint.}
    \label{fig:fig_pamg_ablation_vis}
\end{figure}

To verify the necessity of each component in the proposed energy function, we conduct component-wise ablation experiments on the SIRSTD-Pixel dataset. Each variant removes one term from the full PAMG, including the size prior, the saliency term, the homogeneity term, and the geometric prior. The results are reported in Table~\ref{tab:pamg_ablation}.

The full PAMG achieves the highest performance, with an mIoU of 88.34\% and an Area Ratio of 1.22, indicating a reasonable balance between target completeness and scale consistency. This behavior is also evident in Fig.~\ref{fig:fig_pamg_ablation_vis}(b), where the generated mask tightly covers the target while effectively suppressing the clutter region. Removing the size prior reduces the mIoU to 75.68\% and the Area Ratio to 0.86, showing that the growth process tends to stop prematurely without the initial gain term. As shown in Fig.~\ref{fig:fig_pamg_ablation_vis}(c), the resulting mask becomes overly conservative and fails to recover the full target extent.

The most significant degradation occurs when the saliency term is removed, where the mIoU drops to 39.63\% and the Area Ratio increases to 3.29. This result suggests that local contrast is the primary cue for separating targets from background clutter. The visual example in Fig.~\ref{fig:fig_pamg_ablation_vis}(d) further shows that without explicit saliency guidance, PAMG is prone to drifting toward clutter responses. Removing the homogeneity term also causes a notable drop in mIoU (49.26\%), indicating that brightness cues alone are insufficient for stable mask growth. As illustrated in Fig.~\ref{fig:fig_pamg_ablation_vis}(e), the target region can still be roughly localized, but the recovered mask becomes less compact and less faithful to the true target shape.

Removing the geometric prior results in an mIoU of 75.02\%, while the Area Ratio sharply increases to 13.40, indicating severe over-expansion. As illustrated in Fig.~\ref{fig:fig_pamg_ablation_vis}(f), without spatial constraints the growth process easily spreads along structured background regions and absorbs large clutter areas. This result highlights the importance of the geometric prior in constraining spatial propagation and maintaining the scale consistency of the generated masks.

Overall, the quantitative and qualitative results consistently demonstrate that the components of PAMG play complementary roles. The size prior prevents trivial small-area solutions, the saliency term provides the dominant target-clutter contrast cue, the homogeneity term improves structural consistency, and the geometric prior prevents uncontrolled expansion into surrounding clutter. Their combination leads to the most reliable pseudo-mask generation performance.

\begin{table}[t]
\centering
\caption{Component-wise ablation study of PAMG on the SIRSTD-Pixel dataset. Area Ratio denotes the average ratio between the predicted mask area and the ground-truth area.}
\label{tab:pamg_ablation}
\renewcommand{\arraystretch}{1.15}
\begin{tabular}{lcc}
\toprule
\textbf{Variant} & \textbf{mIoU (\%)} & \textbf{Area Ratio} \\
\midrule
Full PAMG & \textbf{88.34} & \textbf{1.22} \\
w/o Size Prior & 75.68 & 0.86 \\
w/o Saliency & 39.63 & 3.29 \\
w/o Homogeneity & 49.26 & 0.88 \\
w/o Geometric Prior & 75.02 & 13.40 \\
\bottomrule
\end{tabular}
\end{table}

\subsubsection{Empirical Validation of the Generation Boundary}

To further examine the practical implications of the boundary condition derived in Sec.~\ref{sec:pamg_method}, we statistically analyze the relationship between the degree to which a target satisfies the theoretical condition and the quality of the generated pseudo masks. Specifically, for each target in the SIRSTD-Pixel dataset, we compute its SCR and the corresponding theoretical boundary value $B(n,\gamma,R_s)$ on the right-hand side of Eq.~(\ref{equ:generation_boundary}). We then define the boundary satisfaction ratio as
\begin{equation}
\rho = \frac{\mathrm{SCR}}{B(n,\gamma,R_s)} .
\end{equation}
When $\rho > 1$, the target satisfies the theoretical generation condition; larger values of $\rho$ indicate that the physical separability of the target is increasingly stronger relative to the theoretical lower bound.

\begin{figure}[t]
    \centering
    \includegraphics[width=0.8\linewidth]{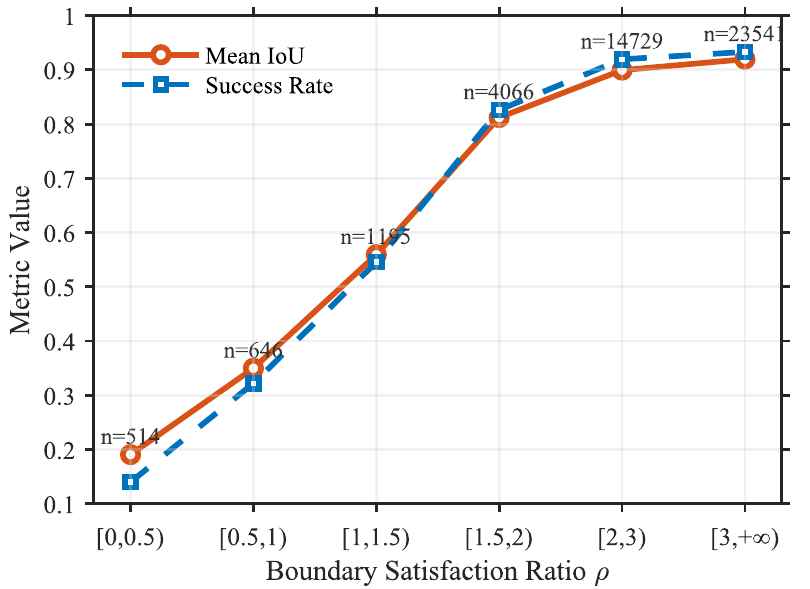}
    \caption{Empirical validation of the theoretical generation boundary. Samples are grouped according to the boundary satisfaction ratio $\rho = \mathrm{SCR}/B(n,\gamma,R_s)$. The average IoU and success rate (IoU $>0.5$) of the pseudo masks generated by PAMG are reported for each interval. As $\rho$ increases, both metrics improve monotonically and gradually stabilize, indicating that targets satisfying the theoretical boundary condition lead to more reliable mask generation.}
    \label{fig:fig_pamgboundary}
\end{figure}

In the experiment, all samples are grouped into several intervals according to their $\rho$ values. For each interval, we compute the average IoU of the pseudo masks generated by PAMG and the success rate (IoU $>0.5$). The statistical results are illustrated in Fig.~\ref{fig:fig_pamgboundary}. As $\rho$ increases, the pseudo-mask quality shows a clear monotonic improvement. When $\rho < 1$, the generation results are relatively unstable, with both the average IoU and success rate remaining low. Once $\rho$ exceeds 1, both metrics increase significantly and gradually stabilize when $\rho > 2$.

This trend is highly consistent with the theoretical analysis. According to Eq.~(\ref{equ:generation_boundary}), when the physical contrast of the target exceeds the derived lower bound, the energy function admits a finite extremum, ensuring that the region-growing process can stably converge near the target support region. The experimental results show that as the degree of boundary satisfaction increases, the consistency between the pseudo masks generated by PAMG and the true target region improves significantly, thereby providing statistical evidence for the validity and practical effectiveness of the derived generation boundary condition.

\subsection{Ablation Studies of RPR-Net}
\label{subsec:ablation}

To rigorously verify the contribution of each component in the proposed RPR-Net and justify the architectural choices, we conducted comprehensive ablation studies on the SIRSTD-Pixel dataset. The results are summarized in Table \ref{tab:ablation}. We analyze the impact from two primary dimensions: temporal modeling strategies and backbone architecture design.

\subsubsection{Impact of Temporal Modeling Strategies}
To demonstrate the necessity of the proposed Temporal Difference Attention and Global Gating mechanism, we compared the full model with three variants:
\begin{itemize}
    \item Single-Frame Baseline (Variant-Single): Removing all temporal modules and inputting only the current frame.
    \item Naive Concatenation (Variant-Concat): Replacing TDA with a simple channel concatenation of the three frames followed by a $1\times1$ convolution.
    \item Without Gating (Variant-NoGate): Using TDA but removing the learnable global gating mechanism (forcing the gate value to 1.0).
\end{itemize}

As shown in Table \ref{tab:ablation}, the \textit{Variant-Single} suffers a clear drop in detection rate ($P_d$ drops by 8.06\%), suggesting that motion cues are important for distinguishing dim small targets from clutter. 
The \textit{Variant-Concat} performs even worse than the single-frame baseline ($P_d$ 78.18\% vs. 83.48\%). This indicates that naive temporal fusion introduces interference due to background misalignment and non-rigid motion, rather than consistently providing useful information. This observation supports the use of differential attention to selectively enhance motion saliency.
Furthermore, the \textit{Variant-NoGate} exhibits a low $P_d$ (84.95\%) and an extremely low $F_a$. This suggests that without the gating mechanism to filter out global camera jitter, the network becomes overly conservative, suppressing potential targets to avoid false alarms caused by background motion. The full model achieves the most balanced trade-off among these variants, indicating that the Global Gating mechanism contributes to more stable detection.

\subsubsection{Impact of Backbone Architecture}
We investigate the trade-off between model capacity, resolution, and performance.
\begin{itemize}
    \item Network Width: We scaled the backbone width using a width multiplier. The results show that a smaller width (0.375) leads to under-fitting due to insufficient feature extraction capacity. Conversely, a larger width (0.75) degrades performance ($P_d$ drops to 90.48\%) and doubles the $F_a$, indicating that an overly large model tends to overfit the noise in small target datasets. The default width (0.5) provides the most balanced result.
    \item Downsampling Stride: Comparing our trimmed backbone (Stride 16) with the standard configuration (Stride 32, \textit{Variant-Stride}), the latter shows a clear decline in $P_d$ (88.69\%). This confirms our hypothesis that excessive downsampling leads to feature collapse for small  targets. 
\end{itemize}

\begin{table*}[t]
\centering
\caption{Ablation studies of RPR-Net on the SIRSTD-Pixel dataset. The ``Key Change'' column describes the modification relative to the baseline model.}
\label{tab:ablation}
\renewcommand{\arraystretch}{1.15}
\begin{tabular}{lccccccc}
\toprule
\multirow{2}{*}{\textbf{Model Variant}} & \multirow{2}{*}{\textbf{Key Change}} 
& \multicolumn{2}{c}{\textbf{Backbone Config}} 
& \multicolumn{3}{c}{\textbf{Detection Metrics}} 
& \multicolumn{1}{c}{\textbf{Shape}} \\

\cmidrule(lr){3-4} 
\cmidrule(lr){5-7}
& & \textbf{Width} & \textbf{Max Stride} 
& \textbf{$P_d$ (\%)}  
& \textbf{$F_a$ ($10^{-5}$)} 
& \textbf{AUC}  
& \textbf{mIoU (\%)} \\ 

\midrule
\multicolumn{8}{c}{\textit{Ablation on Backbone Capacity \& Resolution}} \\ 
\midrule
Variant-Width-S & Small Width & 0.375 & 16 & 89.69 & 6.44 & 0.9485 & 34.63 \\
Variant-Width-L & Large Width & 0.75 & 16 & 90.49 & 12.00 & 0.9524 & 29.90 \\
Variant-Stride & Standard Downsample & 0.5 & 32 & 88.69 & 2.60 & 0.9435 & \textbf{39.27} \\

\midrule
\multicolumn{8}{c}{\textit{Ablation on Temporal Modeling}} \\ 
\midrule
Variant-Single & Single Frame Input & 0.5 & 16 & 83.48 & 4.87 & 0.9174 & 12.67 \\
Variant-Concat & Naive Concatenation & 0.5 & 16 & 78.18 & 15.20 & 0.8909 & 22.46 \\
Variant-NoGate & TDA w/o Gating & 0.5 & 16 & 84.95 & \textbf{0.82} & 0.9248 & 21.32 \\

\midrule
RPR-Net (Ours) & Baseline & 0.5 & 16 & \textbf{91.54} & 5.41 & \textbf{0.9577} & 35.44 \\

\bottomrule
\end{tabular}%
\end{table*}

\subsection{Efficiency Analysis}

Table \ref{tab:efficiency} reports the parameter count, FLOPs, and inference latency of different methods on the SIRSTD-Pixel dataset. From the perspective of sequential detection, the proposed RPR-Net achieves the highest AUC of 0.9577 while requiring only 1.55 M parameters and 13.28 G FLOPs, both lower than SSTNet and TMP. Its inference latency is 11.86 ms per sample, which is also the lowest among the compared sequential methods. These results suggest that the proposed geometric regression paradigm can improve detection accuracy while reducing deployment cost.

\begin{table}[ht]
\centering
\caption{Efficiency comparison with state-of-the-art methods on the SIRSTD-Pixel dataset. Parameters, FLOPs, and latency are reported in M, G, and ms, respectively.}
\label{tab:efficiency}

\small
\setlength{\tabcolsep}{3.5pt}

\begin{tabular}{cccccc}
\toprule
\textbf{Type} & \textbf{Method} & \textbf{Params$\downarrow$} & \textbf{FLOPs$\downarrow$} & \textbf{Latency$\downarrow$} & \textbf{AUC$\uparrow$} \\
\midrule

\multirow{9}{*}{\rotatebox{90}{Single-frame}}
& ACM \cite{dai2021asymmetric} & 0.40 & 2.01   & 2.95  & 0.8452 \\
& AGPCNet \cite{zhang2023attentionguided} & 12.36 & 215.91 & 37.75 & 0.8571 \\
& ALCNet \cite{dai2021attentional} & 0.52 & 1.89   & 2.81  & 0.8775 \\
& DNANet \cite{li2022dense} & 4.70 & 71.31  & 79.49 & 0.8390 \\
& ISNet \cite{zhang2022isnet} & 1.09 & 153.09 & 25.60 & 0.8710 \\
& ISTDU-Net \cite{hou2022istdunet} & 2.76 & 39.72  & 13.70 & 0.8577 \\
& MSHNet \cite{liu2024infrared} & 4.07 & 30.53  & 31.44 & 0.8464 \\
& RDIAN \cite{sun2023receptivefield} & 0.22 & 18.59  & 12.31 & 0.8645 \\
& SCTransNet \cite{yuan2024sctransnet} & 11.33 & 50.57 & 59.60 & 0.8626 \\
\midrule

\multirow{5}{*}{\rotatebox{90}{Sequential}}
& DTUM \cite{li2025directioncoded} & 0.30 & 31.75 & 17.01 & 0.8786 \\
& SSTNet \cite{chen2024sstnet} & 11.95 & 61.80 & 19.19 & 0.9166 \\
& TMP \cite{zhu2024tmp} & 16.41 & 46.43 & 22.07 & 0.9260 \\
\cmidrule(lr){2-6}
& RPR-Net & 1.55 & 13.28 & 11.86 & 0.9577 \\
& RPR-Net+PAMG & 1.55 & 13.28$^\dagger$ & 15.78 & 0.9500 \\
\bottomrule
\end{tabular}

\vspace{1mm}
{\footnotesize $^\dagger$ PAMG is a physics-driven algorithm executed on the CPU and introduces no additional network FLOPs.}

\end{table}

The introduction of PAMG does not alter the network complexity of RPR-Net. As shown in Table \ref{tab:efficiency}, RPR-Net+PAMG retains the same learnable parameter count and network FLOPs, while only incurring a modest increase in end-to-end latency due to the additional CPU-side mask recovery procedure. This observation is consistent with the design principle of our framework: computationally intensive dense segmentation is avoided during the main network inference stage, and fine-grained shape recovery is invoked only when needed. Therefore, the proposed framework provides a practical balance between detection efficiency and mask-level perception, making it more suitable for realistic infrared small target applications than end-to-end segmentation pipelines.

\section{Discussion}

\subsection{Applicability of the Overall Framework}

The proposed \textit{Point-to-Mask} framework is particularly suitable for infrared small target detection, where targets often appear as compact thermal responses with weak texture and ambiguous boundaries. In this setting, the most informative cues are usually concentrated in the target location and its effective thermal support region, rather than in highly detailed contours. From this perspective, the proposed pipeline is more compatible with the physical characteristics of infrared small targets than coarse bounding-box representations, which often introduce substantial background redundancy, and may also provide a lighter alternative to dense segmentation models.

From the perspective of PAMG, when the target exhibits sufficient local saliency, relatively homogeneous internal intensity, and a spatially bounded support region, the energy function along the growth trajectory tends to admit a stable optimum, as analyzed in Sec.~\ref{sec:pamg_method}. Under such conditions, sparse point annotations can often be expanded into a compact support region. From the perspective of RPR-Net, the \textit{center--radius} representation is also well matched to the geometry of typical infrared small targets, whose spatial extent is usually limited and structurally simple. In sequential scenes, the temporal difference attention module can further make use of subtle inter-frame variations, which is broadly consistent with the empirical gains observed in the experiments.

\subsection{Boundary of the Formulation}

The performance boundary of the proposed framework is mainly related to the validity of the above physical and geometric assumptions. When these assumptions are weakened, performance degradation becomes more likely.

For PAMG, instability may arise when the target--background contrast becomes extremely weak, when the target interior exhibits strong structural variations, or when nearby bright structures interfere with the growth trajectory. In such cases, the energy formulation may no longer provide a sufficiently clear preference for the true target support region, which can lead to incomplete growth, background leakage, or merged responses.

For RPR-Net, the limitation mainly stems from the expressive capacity of the compact geometric representation. Although the circle-based center--radius formulation is efficient, it still serves as an approximation of the true target support region. When targets exhibit elongated, smeared, or irregular responses, the predicted center and radius may become biased. Moreover, the temporal module is most helpful when inter-frame differences remain informative. Under severe background motion or strong target-like clutter, the benefit of temporal enhancement may be reduced.

Overall, the proposed framework performs best when two conditions hold simultaneously:
(1) point annotations can be reliably converted into compact target support regions, and
(2) compact geometric representations are sufficient to describe the target structure.
When either condition is not well satisfied, the resulting performance degradation may be better understood as a natural boundary of the formulation, rather than as an incidental failure of a particular module.

\section{Conclusion}

This paper presents the Point-to-Mask framework for infrared small target detection by revisiting supervision and target representation from the perspective of infrared imaging characteristics. Different from the prevailing dense prediction paradigm, the proposed framework bridges low-cost point supervision and mask-level detection through a unified pipeline consisting of PAMG and RPR-Net.
Specifically, PAMG recovers compact target masks from arbitrary point annotations and generates geometric supervision in the form of target centroids and effective radii, while RPR-Net utilizes these supervisory signals to reformulate sequential IRSTD as target center localization and effective radius regression with spatiotemporal motion cues.
In addition, we construct the SIRSTD-Pixel dataset to support unified evaluation of point-supervised pseudo-mask generation, mask-level detection, and sequential pixel-level analysis. Extensive experiments demonstrate that the proposed framework achieves strong pseudo-label quality, high detection accuracy, and efficient inference, while maintaining performance close to full supervision under point-supervised settings with substantially lower annotation cost.
These results suggest that recovering compact target support from sparse point annotations and coupling it with efficient geometric regression is a promising direction for IRSTD. Future work will further investigate more complex target morphologies, stronger dynamic background interference, and broader cross-scene generalization.

\appendices
\section{Discrete Energy Increment Analysis}
\label{app:energy_analysis}

This appendix provides a rigorous discrete-scale analysis of the energy
evolution underlying the PAMG framework and establishes the existence of a
finite maximizer along the monotonic region expansion trajectory.

\subsection{Discrete Energy Sequence and Increment}

Let $\{S_n\}_{n\ge1}$ denote the sequence of regions generated by PAMG, where
each expansion step adds exactly one pixel.
The corresponding energy values form a discrete sequence
$\{\mathcal{E}(S_n)\}$.

PAMG does not rely on an online stopping rule.
Instead, all intermediate energy values are recorded and the optimal region is
selected \emph{a posteriori}.
Therefore, the theoretical question reduces to whether
$\{\mathcal{E}(S_n)\}$ admits a finite maximizer.

Since $n$ is discrete, continuous differentiation is not applicable.
We characterize the local energy evolution via the discrete increment
\begin{equation}
\Delta \mathcal{E}(n)
=
\mathcal{E}(S_{n+1}) - \mathcal{E}(S_n).
\end{equation}

\textbf{Proposition 1.}
If there exists a finite integer $n_0$ such that
$\Delta \mathcal{E}(n)<0$ for all $n>n_0$, then
$\mathcal{E}(S_n)$ attains a maximum at some finite
$n^* \le n_0$.

The remainder of this appendix analyzes $\Delta \mathcal{E}(n)$ term by term.

\subsection{Increment of Individual Energy Components}

\paragraph{Dynamic Gain Term.}
The entropy-driven gain term is defined as
\begin{equation}
\mathcal{E}_{gain}(n)=\ln(\ln n).
\end{equation}
Its discrete increment satisfies
\begin{equation}
\Delta \mathcal{E}_{gain}
=
\ln(\ln(n+1))-\ln(\ln n).
\end{equation}
For target regions of typical sizes, the increment can be approximated as
\begin{equation}
\Delta \mathcal{E}_{gain}
\approx
\frac{1}{n\ln n},
\end{equation}
indicating a positive but gradually diminishing growth benefit as the region expands.

\paragraph{Statistical Resistance Term.}
The statistical component of the Hw-SNR energy is
\begin{equation}
\mathcal{E}_{stat}(n)
=
\ln(\mu_{in}-\mu_{out})-\ln(\sigma_{in}+\epsilon).
\end{equation}
The dominant degradation occurs when background pixels are incorporated.
Let $\mu_T$ and $\mu_B$ denote the mean intensities of the target and the local background, respectively, and let $\sigma_T$ and $\sigma_B$ denote their corresponding standard deviations. Let $\sigma_T^2$ further denote the internal variance of the target-support region at scale $n$.
After adding a pixel with intensity deviation
$\Delta\mu=\mu_T-\mu_B$, the variance update admits the conservative bound
\begin{equation}
\sigma_{n+1}^2
\approx
\sigma_T^2+\frac{\Delta\mu^2}{n}.
\end{equation}
The corresponding energy increment is
\begin{equation}
\Delta \mathcal{E}_{stat}
=
-\frac{1}{2}
\ln\!\left(
\frac{\sigma_{n+1}^2}{\sigma_T^2}
\right)
\approx
-\frac{\Delta\mu^2}{2n\sigma_T^2},
\end{equation}
revealing a scale-dependent statistical resistance induced by variance inflation.

\paragraph{Geometric Prior Term.}
The geometric regularization term is
\begin{equation}
\mathcal{E}_{geo}(n)
=
-\frac{d_{max}^2}{2R_s^2}.
\end{equation}
Here, $R_s$ is interpreted as the spatial support scale of the geometric prior, which controls the strength of the distance-based regularization rather than the physical boundary of the target.
Assuming approximately isotropic growth,
$n\approx\pi d_{max}^2$, yielding the constant increment
\begin{equation}
\Delta \mathcal{E}_{geo}
\approx
-\frac{1}{2\pi R_s^2}.
\end{equation}

\subsection{Existence of a Finite Maximizer}

Combining all components, the discrete energy increment admits the approximation
\begin{equation}
\Delta \mathcal{E}(n)
\approx
\frac{1}{n\ln n}
-
\frac{\Delta\mu^2}{2n\sigma_T^2}
-
\frac{1}{2\pi R_s^2}.
\end{equation}
As $n$ increases, the diminishing dynamic gain is eventually dominated by the
statistical and geometric resistance terms, implying
$\Delta \mathcal{E}(n)<0$ beyond a finite scale.
By Proposition~1, $\mathcal{E}(S_n)$ therefore possesses a finite maximizer
along the expansion trajectory.

\subsection{Generation Boundary}

The sufficient condition $\Delta \mathcal{E}(n)<0$ can be written as
\begin{equation}
\frac{\Delta\mu^2}{2n\sigma_T^2}
+
\frac{1}{2\pi R_s^2}
>
\frac{1}{n\ln n}.
\end{equation}
Introducing the SCR
$\mathrm{SCR}=\Delta\mu/\sigma_B$
and the uniformity ratio
$\gamma=\sigma_B/\sigma_T$,
we obtain the generation boundary
\begin{equation}
\mathrm{SCR}
>
\frac{1}{\gamma}
\sqrt{
2n
\max\!\left(
0,
\frac{1}{n\ln n}
-
\frac{1}{2\pi R_s^2}
\right)
}.
\end{equation}

This boundary follows directly from discrete energy increment analysis and
constitutes a sufficient condition for the existence of a finite energy
maximizer, without assuming any online early stopping strategy.

\section{List of Abbreviations}
\label{app:abbreviations}

\begin{table}[ht]
\centering
\caption{Main abbreviations used in this paper.}
\label{tab:abbreviations}
\renewcommand{\arraystretch}{1.1}
\setlength{\tabcolsep}{6pt}
\begin{tabular}{ll}
\toprule
\textbf{Abbreviation} & \textbf{Full Expression} \\
\midrule
IRSTD & Infrared Small Target Detection \\
PAMG & Physics-driven Adaptive Mask Generation \\
RPR-Net & Radius-aware Point Regression Network \\
PSF & Point Spread Function \\
MAP & Maximum A Posteriori \\
Hw-SNR & Homogeneity-Weighted Signal-to-Noise Ratio \\
SCR & Signal-to-Clutter Ratio \\
TDA & Temporal Difference Attention \\
SAB & Spatial Attention Block \\
FPN & Feature Pyramid Network \\
MLP & Multi-Layer Perceptron \\
GAP & Global Average Pooling \\
LWIR & Long-Wave Infrared \\
GT & Ground Truth \\
IoU & Intersection over Union \\
mIoU & Mean Intersection over Union \\
AUC & Area Under Curve \\
\bottomrule
\end{tabular}
\end{table}

\bibliographystyle{ieeetrans}
\bibliography{bib/references}

@article{li2022dense,
  title={Dense nested attention network for infrared small target detection},
  author={Li, Boyang and Xiao, Chao and Wang, Longguang and Wang, Yingqian and Lin, Zaiping and Li, Miao and An, Wei and Guo, Yulan},
  journal={IEEE Transactions on Image Processing},
  volume={32},
  pages={1745--1758},
  year={2022},
  publisher={IEEE}
}

@inproceedings{teutsch2010classification,
  title={Classification of small boats in infrared images for maritime surveillance},
  author={Teutsch, Michael and Kr{\"u}ger, Wolfgang},
  booktitle={2010 international WaterSide security conference},
  pages={1--7},
  year={2010},
  organization={IEEE}
}

@inproceedings{zhang2024irsam,
  title={IRSAM: Advancing segment anything model for infrared small target detection},
  author={Zhang, Mingjin and Wang, Yuchun and Guo, Jie and Li, Yunsong and Gao, Xinbo and Zhang, Jing},
  booktitle={European Conference on Computer Vision},
  pages={233--249},
  year={2024},
  organization={Springer}
}

@article{rawat2020review,
  title={Review on recent development in infrared small target detection algorithms},
  author={Rawat, Sur Singh and Verma, Sashi Kant and Kumar, Yatindra},
  journal={Procedia Computer Science},
  volume={167},
  pages={2496--2505},
  year={2020},
  publisher={Elsevier}
}

@ARTICLE{liu2023infrared,
  author={Liu, Fangcen and Gao, Chenqiang and Chen, Fang and Meng, Deyu and Zuo, Wangmeng and Gao, Xinbo},
  journal={IEEE Transactions on Image Processing}, 
  title={Infrared Small and Dim Target Detection With Transformer Under Complex Backgrounds}, 
  year={2023},
  volume={32},
  number={},
  pages={5921-5932},
  doi={10.1109/TIP.2023.3326396}}

@ARTICLE{luo2025spatialtemporal,
  author={Luo, Yuan and Li, Xiaorun and Chen, Shuhan},
  journal={IEEE Transactions on Multimedia}, 
  title={Spatial-Temporal Aware-Based Unsupervised Network for Infrared Small Target Detection}, 
  year={2025},
  volume={27},
  number={},
  pages={4895-4909},
  doi={10.1109/TMM.2025.3543002}}

@article{liu2023singleframe,
  title={Single-frame infrared small target detection by high local variance, low-rank and sparse decomposition},
  author={Liu, Yujia and Liu, Xianyuan and Hao, Xuying and Tang, Wei and Zhang, Sanxing and Lei, Tao},
  journal={IEEE Transactions on Geoscience and Remote Sensing},
  year={2023},
  publisher={IEEE}
}

@article{li2023sparse,
  title={Sparse regularization-based spatial--temporal twist tensor model for infrared small target detection},
  author={Li, Jie and Zhang, Ping and Zhang, Lingyi and Zhang, Zhiyuan},
  journal={IEEE Transactions on Geoscience and Remote Sensing},
  volume={61},
  pages={1--17},
  year={2023},
  publisher={IEEE}
}

@ARTICLE{ding2025improving,
  author={Ding, Hongwei and Huang, Nana and Wu, Yaoxin and Cui, Xiaohui},
  journal={IEEE Transactions on Multimedia},
  title={Improving Infrared Small Target Detection With GAN-Driven Data Augmentation},
  year={2025},
  volume={27},
  number={},
  pages={9516-9531},
  doi={10.1109/TMM.2025.3613079}}

@ARTICLE{guo2026multiscale,
  author={Guo, Xuedong and Deng, Lei and Li, Maoyong and Chen, Zhixiang and Yu, Heng and Chen, Hanrui and Dong, Mingli and Zhu, Lianqing},
  journal={IEEE Transactions on Multimedia},
  title={Multiscale Feature Fusion Spatial-channel Attention Network for Infrared Small Target Segmentation},
  year={2026},
  volume={},
  number={},
  pages={1-15},
  doi={10.1109/TMM.2026.3655470}}

@ARTICLE{lin2024learning,
  author={Lin, Fanzhao and Ge, Shiming and Bao, Kexin and Yan, Chenggang and Zeng, Dan},
  journal={IEEE Transactions on Multimedia},
  title={Learning Shape-Biased Representations for Infrared Small Target Detection},
  year={2024},
  volume={26},
  number={},
  pages={4681-4692},
  doi={10.1109/TMM.2023.3325743}}

@ARTICLE{yang2024eflnet,
  author={Yang, Bo and Zhang, Xinyu and Zhang, Jian and Luo, Jun and Zhou, Mingliang and Pi, Yangjun},
  journal={IEEE Transactions on Geoscience and Remote Sensing}, 
  title={EFLNet: Enhancing Feature Learning Network for Infrared Small Target Detection}, 
  year={2024},
  volume={62},
  number={},
  pages={1-11},
  doi={10.1109/TGRS.2024.3365677}}

@inproceedings{liu2024infrared,
  title={Infrared Small Target Detection with Scale and Location Sensitivity},
  author={Liu, Y. and others},
  booktitle={Proceedings of the IEEE/CVF Conference on Computer Vision and Pattern Recognition (CVPR)},
  year={2024}
}

@inproceedings{zhang2022isnet,
  title={ISNet: Shape matters for infrared small target detection},
  author={Zhang, Mingjin and Zhang, Rui and Yang, Yuxiang and Bai, Haichen and Zhang, Jing and Guo, Jie},
  booktitle={Proceedings of the IEEE/CVF conference on computer vision and pattern recognition},
  pages={877--886},
  year={2022}
}

@misc{hui2019a,
  author       = {Bingwei Hui and Zhiyong Song and Hongqi Fan and Ping Zhong and Weidong Hu and Xiaofeng Zhang and Jianguo Lin and Hongyan Su and Wei Jin and Yongjie Zhang and Yaxi Bai},
  title        = {{A dataset for infrared image dim-small aircraft target detection and tracking under ground / air background}},
  year         = 2019,
  month        = oct,
  publisher    = {Science Data Bank},
  version      = {V1},
  doi          = {10.11922/sciencedb.902},
  url          = {https://doi.org/10.11922/sciencedb.902}
}

@article{zhu2024tmp,
  title={TMP: Temporal motion perception with spatial auxiliary enhancement for moving infrared dim-small target detection},
  author={Zhu, Sicheng and Ji, Luping and Zhu, Jiewen and Chen, Shengjia and Duan, Weiwei},
  journal={Expert Systems with Applications},
  volume={255},
  pages={124731},
  year={2024},
  publisher={Elsevier}
}

@ARTICLE{ni2025pointtopoint,
  author={Ni, Rixiang and Wu, Jing and Qiu, Zhaobing and Chen, Liqiong and Luo, Changhai and Huang, Feng and Liu, Qiujiang and Wang, Binxing and Li, Yunxiang and Li, Youli},
  journal={IEEE Transactions on Geoscience and Remote Sensing}, 
  title={Point-to-Point Regression: Accurate Infrared Small Target Detection With Single-Point Annotation}, 
  year={2025},
  volume={63},
  number={},
  pages={1-19},
  doi={10.1109/TGRS.2025.3554025}}

@ARTICLE{duan2024tripledomain,
  author={Duan, Weiwei and Ji, Luping and Chen, Shengjia and Zhu, Sicheng and Ye, Mao},
  journal={IEEE Transactions on Geoscience and Remote Sensing}, 
  title={Triple-Domain Feature Learning With Frequency-Aware Memory Enhancement for Moving Infrared Small Target Detection}, 
  year={2024},
  volume={62},
  pages={1-14},
  doi={10.1109/TGRS.2024.3452175}
}

@article{he2025hybrid,
  title={Hybrid mask generation for infrared small target detection with single-point supervision},
  author={He, Weijie and Liu, Mushui and Yu, Yunlong},
  journal={Neurocomputing},
  pages={131688},
  year={2025},
  publisher={Elsevier}
}

@inproceedings{li2023monte,
  title={Monte Carlo linear clustering with single-point supervision is enough for infrared small target detection},
  author={Li, Boyang and Wang, Yingqian and Wang, Longguang and Zhang, Fei and Liu, Ting and Lin, Zaiping and An, Wei and Guo, Yulan},
  booktitle={Proceedings of the IEEE/CVF international conference on computer vision},
  pages={1009--1019},
  year={2023}
}

@inproceedings{kirillov2023segment,
  title={Segment anything},
  author={Kirillov, Alexander and Mintun, Eric and Ravi, Nikhila and Mao, Hanzi and Rolland, Chloe and Gustafson, Laura and Xiao, Tete and Whitehead, Spencer and Berg, Alexander C and Lo, Wan-Yen and others},
  booktitle={Proceedings of the IEEE/CVF international conference on computer vision},
  pages={4015--4026},
  year={2023}
}

@inproceedings{ying2023mapping,
  title={Mapping degeneration meets label evolution: Learning infrared small target detection with single point supervision},
  author={Ying, Xinyi and Liu, Li and Wang, Yingqian and Li, Ruojing and Chen, Nuo and Lin, Zaiping and Sheng, Weidong and Zhou, Shilin},
  booktitle={Proceedings of the IEEE/CVF Conference on Computer Vision and Pattern Recognition},
  pages={15528--15538},
  year={2023}
}

@inproceedings{yu2025from,
  title={From easy to hard: Progressive active learning framework for infrared small target detection with single point supervision},
  author={Yu, Chuang and Zhao, Jinmiao and Liu, Yunpeng and Zhao, Sicheng and Dai, Yimian and Yue, Xiangyu},
  booktitle={Proceedings of the IEEE/CVF International Conference on Computer Vision},
  pages={2588--2598},
  year={2025}
}

@article{kou2024mcgc,
  title={MCGC: A multiscale chain growth clustering algorithm for generating infrared small target mask under single-point supervision},
  author={Kou, Renke and Wang, Chunping and Fu, Qiang and Li, Zhanwu and Luo, Ying and Li, Boyang and Li, Wei and Peng, Zhenming},
  journal={IEEE Transactions on Geoscience and Remote Sensing},
  volume={62},
  pages={1--12},
  year={2024},
  publisher={IEEE}
}

@ARTICLE{yue2025sdsnet,
  author={Yue, Taoran and Lu, Xiaojin and Cai, Jiaxi and Chen, Yuanping and Chu, Shibing},
  journal={IEEE Transactions on Geoscience and Remote Sensing}, 
  title={SDS-Net: Shallow–Deep Synergism-Detection Network for Infrared Small Target Detection}, 
  year={2025},
  volume={63},
  number={},
  pages={1-13},
  doi={10.1109/TGRS.2025.3588117}}

@ARTICLE{tong2024sttrans,
  author={Tong, Xiaozhong and Zuo, Zhen and Su, Shaojing and Wei, Junyu and Sun, Xiaoyong and Wu, Peng and Zhao, Zongqing},
  journal={IEEE Transactions on Geoscience and Remote Sensing}, 
  title={ST-Trans: Spatial-Temporal Transformer for Infrared Small Target Detection in Sequential Images}, 
  year={2024},
  volume={62},
  pages={1-19},
  doi={10.1109/TGRS.2024.3355947}
}

@ARTICLE{huang2024lmaformer,
  author={Huang, Yuanxin and Zhi, Xiyang and Hu, Jianming and Yu, Lijian and Han, Qichao and Chen, Wenbin and Zhang, Wei},
  journal={IEEE Transactions on Geoscience and Remote Sensing}, 
  title={LMAFormer: Local Motion Aware Transformer for Small Moving Infrared Target Detection}, 
  year={2024},
  volume={62},
  number={},
  pages={1-17},
  doi={10.1109/TGRS.2024.3502663}}

@article{peng2025moving,
  title={Moving infrared dim and small target detection by mixed spatio-temporal encoding},
  author={Peng, Shuang and Ji, Luping and Chen, Shengjia and Duan, Weiwei and Zhu, Sicheng},
  journal={Engineering Applications of Artificial Intelligence},
  volume={144},
  pages={110100},
  year={2025},
  publisher={Elsevier}
}

@article{li2025dbmstn,
  title={DBMSTN: A Dual Branch Multiscale Spatio-Temporal Network for dim-small target detection in infrared image},
  author={Li, Na and Yang, Xiangyu and Zhao, Huijie},
  journal={Pattern Recognition},
  pages={111372},
  year={2025},
  publisher={Elsevier}
}

@ARTICLE{chen2024sstnet,
  author={Chen, Shengjia and Ji, Luping and Zhu, Jiewen and Ye, Mao and Yao, Xiaoyong},
  journal={IEEE Transactions on Geoscience and Remote Sensing}, 
  title={SSTNet: Sliced Spatio-Temporal Network With Cross-Slice ConvLSTM for Moving Infrared Dim-Small Target Detection}, 
  year={2024},
  volume={62},
  pages={1-12},
  doi={10.1109/TGRS.2024.3350024}
}

@ARTICLE{sun2023receptivefield,
  author={Sun, Heng and Bai, Junxiang and Yang, Fan and Bai, Xiangzhi},
  journal={IEEE Transactions on Geoscience and Remote Sensing}, 
  title={Receptive-Field and Direction Induced Attention Network for Infrared Dim Small Target Detection With a Large-Scale Dataset IRDST}, 
  year={2023},
  volume={61},
  pages={1-13},
  doi={10.1109/TGRS.2023.3235150}
}

@ARTICLE{jiang2023antiuav,
  author={Jiang, Nan and Wang, Kuiran and Peng, Xiaoke and Yu, Xuehui and Wang, Qiang and Xing, Junliang and Li, Guorong and Guo, Guodong and Ye, Qixiang and Jiao, Jianbin and Zhao, Jian and Han, Zhenjun},
  journal={IEEE Transactions on Multimedia}, 
  title={Anti-UAV: A Large-Scale Benchmark for Vision-Based UAV Tracking}, 
  year={2023},
  volume={25},
  pages={486-500},
  doi={10.1109/TMM.2021.3128047}
}

@article{gao2013infrared,
  title={Infrared patch-image model for small target detection in a single image},
  author={Gao, Chenqiang and Meng, Deyu and Yang, Yi and Wang, Yongtao and Zhou, Xiaofang and Hauptmann, Alexander G},
  journal={IEEE transactions on image processing},
  volume={22},
  number={12},
  pages={4996--5009},
  year={2013},
  publisher={IEEE}
}

@article{chen2013a,
  title={A local contrast method for small infrared target detection},
  author={Chen, CL Philip and Li, Hong and Wei, Yantao and Xia, Tian and Tang, Yuan Yan},
  journal={IEEE transactions on geoscience and remote sensing},
  volume={52},
  number={1},
  pages={574--581},
  year={2013},
  publisher={IEEE}
}

@inproceedings{dai2021asymmetric,
  author    = {Dai, Yimian and Wu, Yiquan and Zhou, Fei and Barnard, Kobus},
  title     = {Asymmetric Contextual Modulation for Infrared Small Target Detection},
  booktitle = {Proceedings of the IEEE/CVF Winter Conference on Applications of Computer Vision},
  pages     = {950--959},
  year      = {2021}
}

@article{dai2021attentional,
  author  = {Dai, Yimian and Wu, Yiquan and Zhou, Fei and Barnard, Kobus},
  title   = {Attentional Local Contrast Networks for Infrared Small Target Detection},
  journal = {IEEE Transactions on Geoscience and Remote Sensing},
  volume  = {59},
  number  = {11},
  pages   = {9813--9824},
  year    = {2021},
  doi     = {10.1109/TGRS.2020.3044958}
}

@article{zhang2023attentionguided,
  author    = {Zhang, Tianfang and Li, Lei and Cao, Siying and Pu, Tian and Peng, Zhenming},
  title     = {Attention-Guided Pyramid Context Networks for Detecting Infrared Small Target under Complex Background},
  journal   = {IEEE Transactions on Aerospace and Electronic Systems},
  year      = {2023},
  doi       = {10.1109/TAES.2023.3238703},
  publisher = {IEEE}
}

@article{hou2022istdunet,
  author  = {Hou, Qingyu and Zhang, Lijun and Tan, Feng and Xi, Yong and Zheng, Haotian and Li, Ning},
  title   = {ISTDU-Net: Infrared Small-Target Detection U-Net},
  journal = {IEEE Geoscience and Remote Sensing Letters},
  volume  = {19},
  pages   = {1--5},
  year    = {2022},
  doi     = {10.1109/LGRS.2022.3141584}
}

@article{yuan2024sctransnet,
  author  = {Yuan, Shuai and Qin, Hanlin and Yan, Xiang and Akhtar, Naveed and Mian, Ajmal},
  title   = {SCTransNet: Spatial-Channel Cross Transformer Network for Infrared Small Target Detection},
  journal = {IEEE Transactions on Geoscience and Remote Sensing},
  volume  = {62},
  pages   = {1--15},
  year    = {2024},
  doi     = {10.1109/TGRS.2024.3383649}
}

@article{li2024a,
  author  = {Li, Haoqing and Yang, Jinfu and Xu, Yifei and Wang, Runshi},
  title   = {A Level Set Annotation Framework With Single-Point Supervision for Infrared Small Target Detection},
  journal = {IEEE Signal Processing Letters},
  volume  = {31},
  pages   = {451--455},
  year    = {2024},
  doi     = {10.1109/LSP.2024.3356411}
}

@article{li2025directioncoded,
  author  = {Li, Ruojing and An, Wei and Xiao, Chao and Li, Boyang and Wang, Yingqian and Li, Miao and Guo, Yao},
  title   = {Direction-Coded Temporal U-Shape Module for Multiframe Infrared Small Target Detection},
  journal = {IEEE Transactions on Neural Networks and Learning Systems},
  volume  = {36},
  number  = {1},
  pages   = {555--568},
  year    = {2025},
  doi     = {10.1109/TNNLS.2023.3331004}
}

\end{document}